\documentclass[11pt]{article}

\usepackage[final]{acl}

\usepackage{microtype}
\usepackage{amsmath}
\usepackage{amsfonts}
\usepackage{booktabs}
\usepackage{graphicx}
\usepackage{xcolor}
\usepackage{multirow}
\usepackage[table]{xcolor}
\usepackage[most]{tcolorbox}
\usepackage{bm}

\newtcolorbox[auto counter, number within=section]{casebox}[1][]{
  breakable,
  colback=white,
  colframe=black,
  boxrule=0.4pt,
  left=4pt, right=4pt, top=4pt, bottom=4pt,
  title={Case~\thetcbcounter: #1}
}

\newcommand{\E}{\mathbb{E}}
\newcommand{\Var}{\mathrm{Var}}
\newcommand{\Lau}{\mathcal L_{\mathrm{AU}}}   
\newcommand{\Lorig}{\mathcal L_{0}}          
\newcommand{\Lstar}{\mathcal L^\star}        

\usepackage[english,bidi=default]{babel} 
\babelfont{rm}{TeXGyreTermesX} 
\babelprovide[import]{hindi}
\babelfont[*devanagari]{rm}{Lohit Devanagari}
\babelprovide[import]{arabic}
\babelfont[*arabic]{rm}{Noto Sans Arabic}

\title{SDAR-VL: Stable and Efficient Block-wise Diffusion for Vision-Language Understanding}

\author{
Shuang Cheng$^{1,2\dagger}$
\quad Yuhua Jiang$^3$
\quad Zineng Zhou$^5$
\quad Dawei Liu$^{2,4}$
\quad Wang Tao$^{5}$\\
\bf{
Linfeng Zhang$^{4}$
\quad Biqing Qi$^2$\thanks{~Corresponding Authors.}
\quad Bowen Zhou$^2$\footnotemark[1]}\\
$^1$Zhejiang University\quad
$^2$Shanghai AI Laboratory \quad $^3$Tsinghua University \\
$^4$Shanghai Jiao Tong University \quad $^5$ByteDance\\
\texttt{\{chengshuang, qibiqing\}@pjlab.org.cn}
}

\begin{document}

\maketitle
\begingroup
\renewcommand\thefootnote{}\footnotetext{
Code: \url{https://github.com/JetAstra/SDAR-VL}
}
\endgroup
\begingroup
\renewcommand\thefootnote{}\footnotetext{
Models: \url{https://huggingface.co/collections/JetLM/sdar-vl}
}
\endgroup
\begin{abstract}
Block-wise discrete diffusion offers an attractive balance between parallel generation and causal dependency modeling, making it a promising backbone for vision-language modeling. However, its practical adoption has been limited by high training cost, slow convergence, and instability, which have so far kept it behind strong autoregressive (AR) baselines.
We present \textbf{SDAR-VL}, the first systematic application of block-wise discrete diffusion to large-scale vision-language understanding (VLU), together with an \emph{integrated framework for efficient and stable training}. This framework unifies three components: (1) \textbf{Asynchronous Block-wise Noise Scheduling} to diversify supervision within each batch; (2) \textbf{Effective Mask Ratio Scaling} for unbiased loss normalization under stochastic masking; and (3) a \textbf{Progressive Beta Noise Curriculum} that increases effective mask coverage while preserving corruption diversity.
Experiments on 21 single-image, multi-image, and video benchmarks show that SDAR-VL consistently improves \emph{training efficiency}, \emph{convergence stability}, and \emph{task performance} over conventional block diffusion. On this evaluation suite, SDAR-VL sets a new state of the art among diffusion-based vision-language models and, under matched settings, matches or surpasses strong AR baselines such as LLaVA-OneVision as well as the global diffusion baseline LLaDA-V, establishing block-wise diffusion as a practical backbone for VLU.
\end{abstract}

\section{Introduction}
\label{sec:introduction}

Vision-Language Models (VLMs) are becoming the backbone of multimodal AI, and robust multimodal \emph{understanding} is a necessary foundation for future unified systems that can both understand and generate visual--textual content~\cite{comanici2025gemini,hurst2024gpt}. 
Current state-of-the-art vision-language understanding models are dominated by autoregressive (AR) decoders such as Qwen-VL, InternVL, and the LLaVA family~\cite{bai2025qwen2,wang2025internvl3,li2024llava}. 
Their left-to-right factorization is well aligned with text generation, but also enforces a strict causal order that limits parallelism and makes it difficult to support the flexible decoding patterns desired in multimodal generation~\cite{xin2025lumina}.

Discrete diffusion models~\cite{austin2021structured,gulrajani2023likelihood} offer an alternative sequence modeling paradigm: sequences are generated by iterative denoising from corrupted inputs, enabling parallel decoding and bidirectional context. 
They have already shown strong performance in multimodal \emph{generation} tasks such as image synthesis and infilling~\cite{xin2025lumina,li2025lavidao,yang2025mmada}, and are therefore promising as a unified backbone for both understanding and generation~\cite{zhu2025llada,ye2025dream}. 
However, existing \emph{global} discrete diffusion approaches face two key obstacles when scaled to vision-language understanding: (1) full-sequence bidirectional denoising leads to per-step quadratic complexity and low throughput, motivating KV-cache–style accelerations~\cite{peng2025efficient,wu2025fast,ma2025dkv}; and (2) the lack of causal inductive bias makes long-form text generation difficult, so practical systems often fall back to semi-autoregressive schemes~\cite{zhu2025llada,yang2025diffusion}. 
These adaptations implicitly approximate block-wise causal processing, suggesting that a more principled block-wise diffusion formulation could be better suited to large-scale multimodal understanding.

Block-wise diffusion models~\cite{arriola2025block,cheng2025sdar,wang2025diffusion} precisely follow this idea by partitioning sequences into autoregressive blocks, while performing masked diffusion in parallel \emph{within} each block. 
This preserves inter-block causal dependencies, reduces the computational cost of global diffusion, and still benefits from local bidirectional context. 
Despite encouraging results in pure-language settings~\cite{cheng2025sdar}, it remains unclear whether block-wise discrete diffusion can serve as a competitive and scalable backbone for vision-language understanding.

In this work, we introduce \textbf{SDAR-VL}, the first block-wise discrete diffusion framework for large-scale vision-language understanding. 
Built on the Block Discrete Denoising Diffusion (BD3) formulation~\cite{arriola2025block,cheng2025sdar}, SDAR-VL targets the {understanding} side and addresses three BD3 training bottlenecks---high gradient variance, biased loss scaling under stochastic masking, and suboptimal noise scheduling---via an integrated framework (Sec.~\ref{sec:method}): 
(1) \textbf{Asynchronous Block-wise Noise Schedule} to smooth optimization by sampling noise per block; 
(2) \textbf{Effective Mask Ratio Scaling} to normalize loss using the realized mask ratio; and 
(3) \textbf{Progressive Beta Noise Curriculum} to shift toward higher mask ratios while preserving corruption diversity. 

We instantiate SDAR-VL with 4B- and 8B-parameter models coupling a block-diffusion language backbone with a SigLIP-2 vision encoder and train them under a four-stage curriculum that progressively introduces alignment, capability expansion, reasoning, and long chain-of-thought (CoT) signals. 
Across 21 single-image, multi-image, and video benchmarks, SDAR-VL achieves state-of-the-art performance among diffusion-based multimodal models and, under matched vision towers, data, and training budgets, consistently outperforms the best global diffusion baseline LLaDA-V~\cite{you2025llada}. 
Moreover, under comparable settings, SDAR-VL matches or surpasses the autoregressive LLaVA-OneVision~\cite{li2024llava}, and long CoT distillation further strengthens performance on challenging multimodal reasoning benchmarks.
In summary, our main contributions are:
\begin{itemize}
    \item We provide the first systematic study of blockwise discrete diffusion for vision-language understanding, showing it matches or surpasses strong autoregressive VLMs while retaining diffusion's structural flexibility.
    \item We propose an integrated training framework---asynchronous block-wise noise scheduling, effective mask ratio scaling, and a progressive Beta noise curriculum---that jointly improves stability and efficiency for large-scale block-wise diffusion.
    \item Through extensive experiments on 21 benchmarks, we demonstrate that SDAR-VL sets a new state of the art among diffusion-based vision-language models and, under matched conditions, consistently outperforms both the global diffusion baseline LLaDA-V and the autoregressive LLaVA-OneVision. With long chain-of-thought distillation, the SDAR-VL-Think variants further reach parity with or surpass CoT-enhanced autoregressive baselines such as the Qwen2.5-VL–based R1-OneVision on math benchmarks.
\end{itemize}

\section{Preliminary}
\label{sec:preliminaries}

We build our work on the Block Discrete Denoising Diffusion framework (BD3)~\cite{arriola2025block,cheng2025sdar}, which interpolates between autoregressive and diffusion paradigms. A token sequence $\mathbf{x}$ of length $L$ is partitioned into $B$ non-overlapping blocks $\{x^1, \dots, x^B\}$, each of length $L' = L/B$. The model factorizes the likelihood autoregressively over blocks:
\begin{equation}
\log p_\theta (\mathbf{x}) = \sum_{b=1}^B \log p_\theta (x^b \mid x^{<b}),
\label{eq:bd3_factorization}
\end{equation}
where $x^{<b}$ denotes all preceding clean blocks. Each conditional distribution $p_\theta (x^b \mid x^{<b})$ is modeled via a discrete diffusion process: a forward process gradually masks tokens in $x^b$, and a reverse process, parameterized by Transformer with block-causal attention, reconstructs the original block.

The standard BD3 training objective applies the Negative Evidence Lower Bound (NELBO) to each block-conditional distribution, scaling the per-block loss by the inverse of the noise ratio $t$:
\begin{equation}
\mathcal{L}(\theta) 
= \mathbb{E}_{x, b, t} \left[
\frac{-1}{t}
\sum_{\ell \in \mathcal{M}_t^b}
\log p_\theta\big(x_0^{b,\ell} \mid x_t^b, x^{<b}\big)
\right]
\label{eq:bd3_loss}
\end{equation}
where $\mathcal{M}_t^b$ is the set of masked positions in block $b$ at noise level $t$. 
Here, $1/t$ compensates for reduced contributing tokens, but—as we discuss in Sec.~\ref{sec:effective-mask-scaler}—this practice can introduce bias when $t$ and the realized mask proportion differ.

\section{Method: The SDAR-VL Framework}
\label{sec:method}

SDAR-VL augments the BD3 training paradigm to reduce gradient variance and improve convergence under fluctuating corruption difficulty. It consists of three components: (1) \textbf{Asynchronous Block-wise Noise Scheduling} (ABNS) to diversify within-step training signals and stabilize optimization, (2) \textbf{Effective Mask Ratio Scaling} (EMRS) to provide unbiased loss normalization, and (3) \textbf{Progressive Beta Distribution Noise Curriculum} (PBNC) to balance noise coverage and diversity, yielding a more stable and efficient block-wise discrete diffusion training framework.

\subsection{Asynchronous Block-wise Noise Schedule}
\label{sec:asynchronous-noise-scheduler}

The conventional BD3 framework uses a \emph{synchronous} noise schedule: a single noise level \(t\) is sampled and applied to all blocks in a training sample, so every block experiences the same masking ratio. As shown in Sec.~\ref{sec:ablations} (Fig.~\ref{fig:relation_t_loss}), the average loss increases monotonically with \(t\), so sampling one \(t\) per step introduces large step-to-step difficulty fluctuations and high gradient variance.

We propose an \emph{asynchronous} block-wise noise schedule, where each block \(b\) independently samples its corruption level:
\begin{equation}
t_b \sim \mathcal{P}(\cdot \mid \tau), \quad b=1,\dots,B.
\end{equation}
Let
\begin{equation}
\ell_b = \sum_{\ell \in \mathcal{M}_{t_b}^b} 
    \log p_\theta\big(x_0^{b,\ell} \mid x_{t_b}^b, x^{<b}\big)
\end{equation}
denote the negative log-likelihood for block \(b\). The modified asynchronous training objective is:
\begin{equation}
\mathcal{L}_{\text{async}}(\theta) 
= \mathbb{E}_{x,\, b,\, t_b} 
\left[ -\frac{\ell_b}{t_b} \right].
\end{equation}

Block-specific corruption exposes the model to a richer distribution of reconstruction difficulties within each step, averaging losses over diverse \(t_b\) and smoothing fluctuations from extreme draws. Appendix~\ref{subsec:theory-async-variance} shows that this asynchronous sampling leaves the expected objective unchanged while reducing the variance of the mini-batch estimator, explaining the improved training stability.

\subsection{Effective Mask Ratio for Unbiased Loss Scaling}
\label{sec:effective-mask-scaler}

Under ABNS, the standard BD3 loss scaling multiplies each block's loss by \(1/t_b\), implicitly assuming that the sampled ratio \(t_b\) matches the realized masked proportion in that block. However, discrete stochastic masking often causes deviations between \(t_b\) and the realized ratio, especially for small blocks or high-variance schedules. 
This mismatch yields biased NELBO estimates and noisier scaling; with independent \(t_b\) across blocks, it can further increase gradient variance and offset the stability gains of asynchronous scheduling. 
Appendix~\ref{subsec:theory-bias} shows that the discrepancy grows with the variance of the discrete masking process.

To resolve this issue, we introduce the \textbf{effective mask ratio}:
\begin{equation}
t'_b = \frac{\|\mathbf{m}^b\|_1}{L'},
\end{equation}
where \(\mathbf{m}^b \in \{0,1\}^{L'}\) is the binary mask vector for block~\(b\), and \(L'\) is the block length. 
By replacing the sampled \(t_b\) with the realized \(t'_b\) in the normalization term, we obtain an \emph{unbiased} scaling:
\begin{equation}
\mathcal{L}_{\text{async+unbiased}}(\theta) 
= \mathbb{E}_{x,\, b,\, t_b} 
\left[ -\frac{\ell_b}{t'_b} \right],
\end{equation}

Intuitively, this simply weights each block by its \emph{realized} corruption level rather than the nominal target. Appendix~\ref{subsec:theory-bias} shows that the resulting \(1/t'_b\) scaling is unbiased for the ideal objective (unlike the standard \(1/t_b\) scaling), and empirically it removes systematic mis-scaling, lowers gradient variance, and stabilizes training.

\subsection{Progressive Beta Distribution Noise Curriculum}
\label{sec:progressive-beta}

A key lever to accelerate diffusion training is to \emph{increase the effective mask ratio} so that more tokens contribute to the loss in each step, improving sample efficiency~\cite{arriola2025block}. 
However, naively shifting all sampled noise ratios toward high values risks collapsing noise diversity, reducing the model's exposure to varied reconstruction difficulties and potentially harming generalization.

\begin{figure}[htbp]
    \centering
    \includegraphics[width=0.46\textwidth]{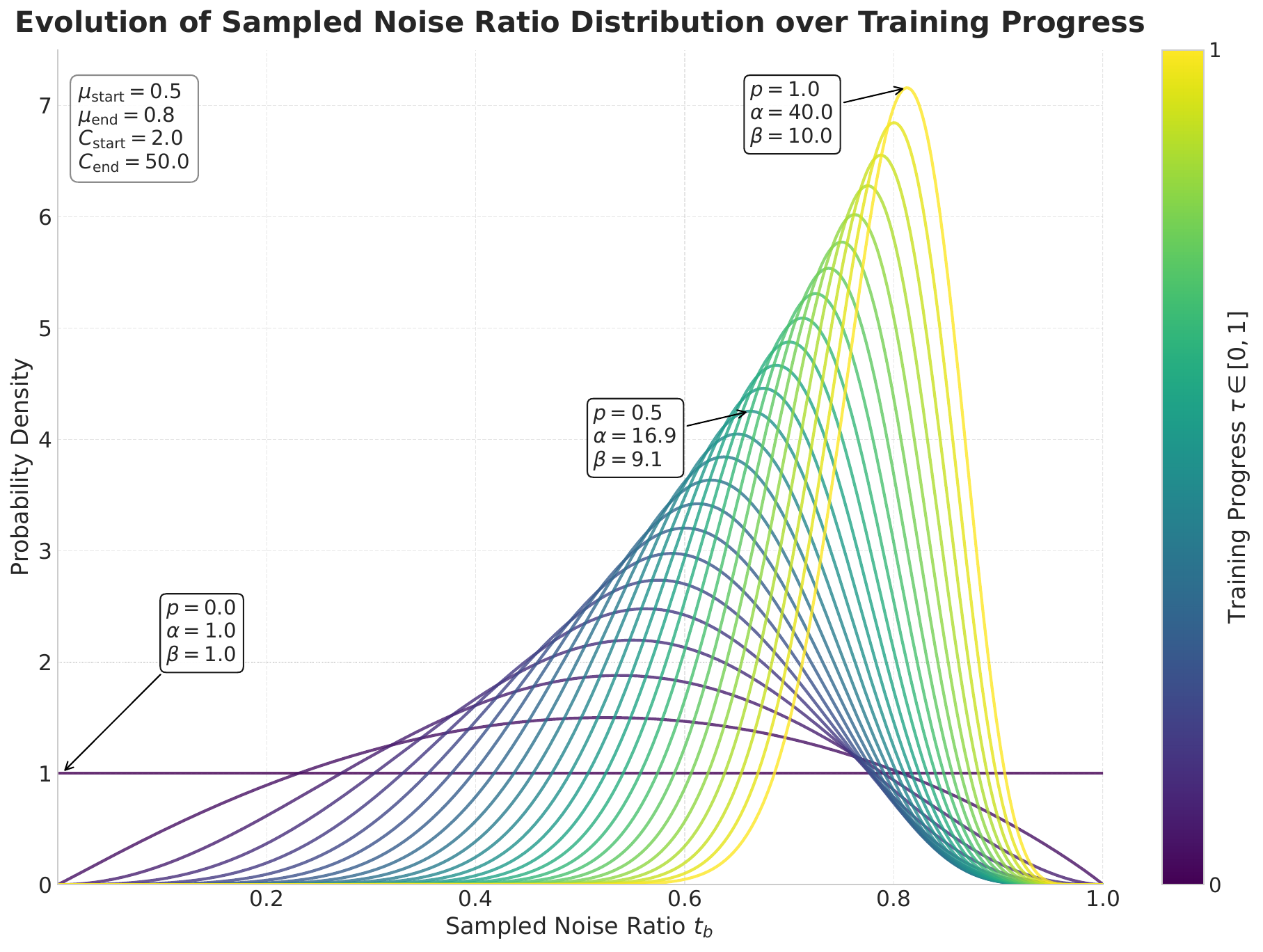}
    \caption{\textbf{Evolution of the progressive Beta noise scheduler.}
    Probability density functions of the block-level noise ratio $t_b$ at different normalized training progress values $u_\tau$ (color-coded). 
    As training progresses, the distribution shifts and sharpens toward higher $t_b$, increasing masked tokens per sample while preserving corruption diversity.}
    \label{fig:dynamic_beta_scheduler}
\end{figure}

We address this trade-off with a progressive Beta scheduler for $\mathcal{P}(t_b \mid \tau)$, which preserves a wide support over $t_b$ while smoothly increasing the mean $\mu_\tau$ and concentration $C_\tau$:
\begin{align}
    &t_b \sim \mathrm{Beta}(\alpha_\tau, \beta_\tau)\\
    &\mu_\tau = \frac{\alpha_\tau}{\alpha_\tau+\beta_\tau},\qquad
    C_\tau = \alpha_\tau+\beta_\tau.
\end{align}
We start from a moderate mean $\mu_\tau$ and low concentration $C_\tau$, yielding a broad distribution over mid-range corruption levels for stable early training. 
Over a warmup horizon, both parameters increase: the mean raises the expected mask ratio (more supervised tokens per step), while the concentration tightens the distribution around this mean without collapsing its support. 
PBNC thus maintains noise diversity while gradually shifting toward higher-coverage, harder reconstructions. 
Figure~\ref{fig:dynamic_beta_scheduler} visualizes this curriculum from easier, varied tasks to harder, high-coverage ones.

\section{Experiment}

\subsection{Training Setting}

\paragraph{Model Architecture}
Our experiments leverage the {SDAR-Chat model}~\cite{cheng2025sdar}, a novel block diffusion-based large language model, in its 4B- and 8B-parameter variants. For the vision encoder, we utilize SigLIP-2-So400m-patch14-384~\cite{tschannen2025siglip}, renowned for its robust visual representation. The vision and language modalities are bridged by a projector, implemented as a randomly initialized two-layer MLP, to align the feature spaces.

\paragraph{Datasets and Training Stages}
We adopt a four-stage curriculum for vision–language alignment, multimodal capability, and long chain-of-thought (CoT) reasoning. 
Stage~1 trains only the projector on LLaVA-Pretrain~\cite{liu2023visual} for vision–language alignment. 
Stage~2 improves general multimodal ability on the MAmmoTH-VL corpus~\cite{guo2024mammoth} using M-SI followed by M-OV. 
Stage~3 strengthens reasoning on VisualWebInstruct (VW)~\cite{jia2025visualwebinstruct} and a mixture of M-OV and VW. 
Stage~4 distills long CoT from R1-OneVision~\cite{yang2025r1} to obtain SDAR-VL-Think. 
Full dataset statistics, stage-wise recipes, and hyperparameters are provided in Appendix~\ref{app:training_settings} (Table~\ref{tab:training_settings}).

\paragraph{Chain-of-Thought Template Masking}  
To encourage chain-of-thought-style responses, we introduce two special tokens, \texttt{<think>} and \texttt{</think>}, as delimiters of reasoning steps in the output. 
During training on \textit{R1-OneVision}, these tokens are always masked under the masked reconstruction objective, providing direct supervision over reasoning-boundary placement and enabling reliable long-CoT template generation. 
This design improves template adherence, interpretability, and performance on complex multimodal tasks.

\paragraph{Implementation Details}
\label{Implementation_Details}

Our 4B and 8B models share the four-stage pipeline above. 
We apply sequence packing to reduce padding and increase GPU utilization, using boundary-aware attention and loss masks to ensure packed samples remain isolated.
In the initial alignment stage (Stage 1), we exclusively train the projector while keeping both the vision and language towers frozen. In the subsequent fine-tuning stages, we unfreeze all components and perform full-parameter fine-tuning of the vision tower, language tower, and projector, using distinct learning rates to optimize overall performance, as specified in Table~\ref{tab:training_settings}.

\definecolor{controlled}{gray}{0.95}
\begin{table*}[htbp]
\centering
\caption{\textbf{Performance on single-image benchmarks.} Results marked with * are taken from prior work (see Appendix~\ref{appendix:experiment}); all others are from our evaluations. Light-gray rows indicate the controlled group, which shares the same vision tower, image processor, and closely matched training data as our models (external baselines additionally marked with $^\dagger$). Within this group, the best score for each benchmark is in \textbf{bold}.}
\label{tab:image-bench}
\setlength{\tabcolsep}{2pt}

\resizebox{0.95\textwidth}{!}{
\begin{tabular}{@{}l|cccccccc@{}}
\toprule
\multirow{2}{*}{\textbf{Model}} & \textbf{MMBench} & \textbf{SEEDBench} & \textbf{MME} & \textbf{MMVet} & \textbf{MMStar} & \textbf{MMMU} & \textbf{HallBench} & \textbf{MathVista} \\ 
\cmidrule(l){2-9}
& {en-dev} & {image} & {test} & {test} & {test} & {val} & {avg} & {mini} \\ 
\midrule
\multicolumn{9}{c}{\textbf{\emph{Closed-source Models}}} \\
\midrule
GPT-4V* & 75.0 & 49.9 & {517/1409} & 49.9 & 57.1 & 56.8 & - & 49.9 \\ 
GPT-4o* & -    & 76.2 & -          & 76.2 & -    & 69.1 & - & 63.8 \\
\midrule
\multicolumn{9}{c}{\textbf{\emph{Open-source Autoregressive Models}}} \\
\midrule
InternVL2-4B & 77.4 & 73.3 & {1535/531} & 55.7 & 54.2 & 46.1 & 41.2 & 58.6 \\
InternVL2-8B & 82.7 & 75.4 & {1642/577} & 60.8 & 61.2 & 47.9 & 45.3 & 58.2 \\
InternVL2\_5-4B & 82.6 & 74.9 & {1678/633} & 63.3 & 58.5 & 50.2 & 46.8 & 61.6 \\
InternVL2\_5-8B & 84.1 & 77.1 & {1688/643} & 68.1 & 62.6 & 53.4 & 49.5 & 63.2 \\
Qwen2-VL-7B & 81.7 & 76.4 & {1673/641} & 66.5 & 61.1 & 50.4 & 49.4 & 61.3 \\
Qwen2.5-VL-3B & 79.8 & 73.9 & {1564/589} & 66.9 & 56.1 & 49.6 & 44.6 & 61.1 \\ 
Qwen2.5-VL-7B & 84.2 & 77.1 & {1683/624} & 70.7 & 65.4 & 55.1 & 52.3 & 69.7 \\

\rowcolor{controlled}
\textbf{LLaVA-OV-7B$^\dagger$} & 82.1 & \textbf{76.7} & {1570/415} & 58.1 & \textbf{62.0} & 46.4 & 38.1 & {62.0} \\ 

\midrule
\multicolumn{9}{c}{\textbf{\emph{Discrete Diffusion Models}}} \\
\midrule
Dimple-8B*       & 74.6   &  -   & 1514/432 &  -  &  -  & 45.2   &  -    & 42.3 \\ 
LaViDA-L-8B*     & 70.5   &  -   & 1366/341 &  -  &  -  & 43.3   &  -    & 44.8 \\ 
LaViDA-O-8B*     & 76.4   &  -   & 1431/488 &  -  &  -  & 45.1   &  -    & 56.9 \\ 
MMaDA-8B*        & 68.5   &  -   & 1410/242 &  -  &  -  & 30.2   &  -    & 33.7 \\ 

\rowcolor{controlled}
\textbf{LLaDA-V-8B*$^\dagger$} & \textbf{82.9} & 74.8   & {1507/491} & - & 60.1 & 48.6   & -  & 59.7 \\ 

\midrule
\multicolumn{9}{c}{\textbf{\emph{Block-wise Discrete Diffusion Models}}} \\
\midrule

\rowcolor{controlled}
\textbf{SDAR-VL-4B-Inst} & 78.7 & 75.4 & {1562/546} & 59.8 & 56.0 & 49.4 & \textbf{46.8} & 57.6 \\ 

\rowcolor{controlled}
\textbf{SDAR-VL-8B-Inst} & 82.2 & 75.5 & \textbf{1632/569} & \textbf{62.6} & 59.9 & \textbf{53.0} & 44.4 & \textbf{62.5} \\

\bottomrule
\end{tabular}%
}
\resizebox{0.95\textwidth}{!}{
\setlength{\tabcolsep}{2pt}
\begin{tabular}{@{}l|cccccccc@{}}
    \toprule
    \multirow{2}{*}{\textbf{Model}} & \textbf{MathVision} & \textbf{MathVerse} & \textbf{ScienceQA} & \textbf{RealworldQA} & \textbf{InfoVQA} & \textbf{DocVQA} & \textbf{ChartQA} & \textbf{AI2D} \\ 
    \cmidrule(l){2-9}
    & {testmini} & {mini-vo/vd} & {test} & {test} & {val/test} & {val/test} & {test} & {test} \\
    \midrule
    \multicolumn{9}{c}{\textbf{\emph{Closed-source Models}}} \\
    \midrule
    GPT-4V* & - & 32.8/- & 75.7 & 61.4 & - & -/88.4 & 78.5 & 78.2  \\ 
    GPT-4o* & - & 50.2/- & -    & 58.6 & - &-/92.8  & 85.7   & 94.2  \\ 
    \midrule
    \multicolumn{9}{c}{\textbf{\emph{Open-source Autoregressive Models}}} \\
    \midrule
    InternVL2-4B    & 16.7 & 24.7/28.8 & 96.2 & 61.2 & 66.1/- & 88.1/- & 80.9 & 78.9 \\
    InternVL2-8B    & 19.2 & 24.2/32.2 & 97.2 & 64.7 & 73.0/- & 91.0/- & 82.2 & 83.7 \\
    InternVL2\_5-4B & 19.9 & 25.2/35.0 & 97.3 & 64.2 & 72.2/- & 91.1/- & 78.0 & 81.4 \\
    InternVL2\_5-8B & 18.8 & 27.2/39.9 & 98.1 & 69.4 & 75.5/- & 92.0/- & 82.8 & 84.5 \\
    Qwen2-VL-7B     & 18.7 & 25.7/32.9 & 86.1 & 70.4 & 76.3/- & 93.9/- & 83.1 & 83.1 \\
    Qwen2.5-VL-3B   & 21.0 & 35.6/38.6 & 81.7 & 65.2 & 76.1/- & 92.9/- & 84.1 & 81.2 \\
    Qwen2.5-VL-7B   & 25.2 & 42.7/44.6 & 89.3 & 68.5 & 82.2/- & 94.9/- & 86.2 & 84.8 \\

    \rowcolor{controlled}
    \textbf{LLaVA-OV-7B$^\dagger$} & 17.2 & 23.7/31.5 & \textbf{95.3} & \textbf{69.7} & 66.3/- & 87.0/- & 80.2 & \textbf{82.6} \\ 

    \midrule
    \multicolumn{9}{c}{\textbf{\emph{Discrete Diffusion Models}}} \\
    \midrule
    Dimple-8B*   &  -  &  -  & 77.1 &  -  &  -    &  -    & 63.4 & 74.4 \\ 
    LaViDA-8B*   & 20.4 & -/27.2 & 80.2 &  -  & -/34.2 & -/59.0 & 64.6 & 70.0 \\ 
    LaViDA-O-8B* &  -   & -/36.9 & 84.6 &  -  & -/44.6 & -/73.7 & 80.0 & 76.7 \\ 
    MMaDA-8B*    &  -   & -/13.5 & 55.8 &  -  & -/14.9 & -/10.9 &  9.8 & 66.6 \\

    \rowcolor{controlled}
    \textbf{LLaDA-V-8B*$^\dagger$}  &  -  & -/28.5 &  -  & 63.2 & 66.3/- & 83.9/- & 78.3 & 77.8 \\ 

    \midrule
    \multicolumn{9}{c}{\textbf{\emph{Block-wise Discrete Diffusion Models}}} \\
    \midrule

    \rowcolor{controlled}
    \textbf{SDAR-VL-4B-Inst} & 21.3 & 33.5/38.9 & 89.8 & 67.6 & 70.5/- & 86.9/- & 81.8 & 79.1 \\

    \rowcolor{controlled}
    \textbf{SDAR-VL-8B-Inst} & \textbf{24.2} & \textbf{36.5/41.0} & 92.2 & 66.5 & \textbf{73.2/-} & \textbf{88.3/-} & \textbf{82.7} & 79.9 \\

    \bottomrule
\end{tabular}
}
\end{table*}

For the ablation studies presented in Section~\ref{sec:ablations}, we employ a two-stage training protocol. First, we perform vision-language alignment by training only the projector on the LLaVA-Pretrain dataset, identical to Stage 1 of our main pipeline. Second, we conduct full-parameter fine-tuning using the {LLaVA-NeXT} dataset~\cite{liu2024llavanext}. The training hyperparameters are detailed in Table~\ref{tab:training_settings}.

\subsection{Evaluation Setting}

\paragraph{Evaluation Benchmarks}
We evaluate SDAR-VL-4B/8B on 21 benchmarks spanning four multimodal ability categories. 
(1) \emph{General visual understanding and reliability}: core perception, factuality, and robustness on MMBench~\cite{liu2024mmbench}, SEEDBench~\cite{li2023seed}, MME~\cite{fu2025mmecomprehensiveevaluationbenchmark}, MM-Vet~\cite{yu2023mm}, MMStar~\cite{chen2024we}, MMMU~\cite{yue2024mmmu}, and HallusionBench~\cite{guan2024hallusionbench}. 
(2) \emph{Mathematical and scientific reasoning}: visual math and science reasoning on MathVista~\cite{lu2023mathvista}, MathVerse~\cite{zhang2024mathverse}, MathVision~\cite{wang2024measuring}, and ScienceQA~\cite{lu2022learn}. 
(3) \emph{Document-centric VQA}: text-rich understanding on DocVQA~\cite{mathew2021docvqa}, ChartQA~\cite{masry2022chartqa}, InfoVQA~\cite{mathew2022infographicvqa}, and AI2D~\cite{kembhavi2016diagram}. 
(4) \emph{Multi-image and video understanding}: cross-image and temporal reasoning on MuirBench~\cite{wang2024muirbench}, BLINK~\cite{fu2024blink}, MVBench~\cite{li2024mvbench}, VideoMME~\cite{fu2025video}, and MLVU~\cite{zhou2024mlvu}. 
Details of the evaluation setup, metrics, and prompts are provided in Appendix~\ref{appendix:experiment}.

\paragraph{Baselines}
We primarily compare to autoregressive vision–language models trained under settings closely aligned with ours. Our training uses the MAmmoTH~\cite{guo2024mammoth} dataset, which also underlies LLaVA-OneVision~\cite{li2024llava}, making model architecture, base language model, vision encoder, and training budget broadly comparable. 
Accordingly, our main baselines are models of similar scale and configuration, including InternVL-2~\cite{opengvlab2024internvl2}, Qwen2-VL~\cite{wang2024qwen2}, and LLaVA-OneVision-OV~\cite{li2024llava}, as well as newer iterations Qwen2.5-VL~\cite{bai2025qwen2} and InternVL-2.5~\cite{chen2024expanding} to track progress across model generations.

We also compare against diffusion-based language models to contrast learning paradigms, including Dimple~\cite{yu2025dimple}, LLaDA-V~\cite{you2025llada}, LaViDA~\cite{li2025lavida}, LaViDA-O~\cite{li2025lavidao}, and MMaDA~\cite{yang2025mmada}. 
Among these, LLaDA-V uses the same vision tower and training data configuration as our models, enabling a controlled comparison that isolates model design and optimization strategies.

\begin{table}[htbp]
\centering
\caption{\textbf{Performance on multi-image and video benchmarks.} Same notations as in Table~\ref{tab:image-bench}. Rows shaded in light gray indicate the controlled comparison group, and the best results within this group are bold.}
\label{tab:multi-video-bench}
\setlength{\tabcolsep}{2pt}
\resizebox{\columnwidth}{!}{%
\begin{tabular}{l|ccccc}
\toprule
\multirow{2}{*}{\textbf{Model}} & \textbf{MuirB} & \textbf{BLINK} & \textbf{VideoMME} & \textbf{LVBench} & \textbf{MLVU} \\
\cmidrule(l){2-6}
& multi-img &  multi-img & wo-subs & val & m-avg \\
\midrule
GPT-4V*    & 62.3 & 51.1 & 59.9 & 61.3 & 49.2  \\
GPT-4o*    & -    & -    & 71.9 & 66.7 & 64.6  \\
\midrule
InternVL-2-4B    & 40.3 & 47.4 & 53.6 & 51.0 & 58.0  \\
InternVL-2-8B    & 48.5 & 50.3 & 56.1 & 53.6 & 61.5  \\
InternVL-2.5-4B  & 45.1 & 50.7 & 61.6 & 54.2 & 66.7  \\
InternVL-2.5-8B  & 51.2 & 54.8 & 63.8 & 59.5 & 67.2  \\
Qwen2-VL-7B      & 39.9 & 52.9 & 59.7 & 55.1 & 64.3  \\
Qwen2.5-VL-3B    & 46.5 & 48.8 & 58.4 & 52.7 & 63.7 \\
Qwen2.5-VL-7B    & 58.2 & 55.6 & 61.9 & 59.7 & 64.3  \\
\rowcolor{controlled}
\textbf{LLaVA-OV-7B$^\dagger$}      & 40.5 & 52.4 & 58.3 & 57.9 & \textbf{66.8} \\
\midrule
\rowcolor{controlled}
\textbf{LLaDA-V-8B*$^\dagger$}      & 48.3 & -    & 56.1 & -    & 59.5  \\
\midrule
\rowcolor{controlled}
\textbf{SDAR-VL-4B-Inst} & 44.8 & {52.8} & 57.3  & 56.2  & 61.7 \\
\rowcolor{controlled}
\textbf{SDAR-VL-8B-Inst} & \textbf{50.2} & \textbf{52.9} & \textbf{60.8}  & \textbf{59.8}  & 65.0 \\
\bottomrule
\end{tabular}
}
\end{table}
\subsection{Main Results}

We present the main evaluation results in Table~\ref{tab:image-bench} and Table~\ref{tab:multi-video-bench}. 
For fair comparisons, we highlight a \textbf{controlled group (shaded in gray)} that shares the same vision tower, image processor, and largely matched training data configuration with our models; the best results within this group are boldfaced. 
Across single-image, multi-image, and video settings, SDAR-VL delivers strong and stable performance, demonstrating the effectiveness of block-wise discrete diffusion for multimodal learning.

\paragraph{Single-image benchmarks.}
As shown in Table~\ref{tab:image-bench}, SDAR-VL achieves strong performance across general visual understanding, hallucination robustness, mathematical/scientific reasoning, and document-centric VQA. 
Under the controlled setting, our models consistently outperform the closely matched diffusion baseline LLaDA-V and lead the group on most benchmarks (14/21), indicating robust perception and reliable reasoning under matched architectures and data conditions. 
Notably, our advantages are more pronounced on reasoning- and text-rich tasks: the 8B model tops the controlled group on MathVista, MathVision, and MathVerse, while also achieving the best results on InfoVQA, DocVQA, and ChartQA. 
Beyond controlled comparisons, SDAR-VL surpasses all discrete diffusion multimodal models and remains competitive with strong autoregressive baselines. 
This competitiveness is notable given our smaller training scale ($\sim$57B tokens) versus reported scales of Qwen2-VL ($\sim$1.4T), Qwen2.5-VL ($\sim$4.1T), and InternVL2.5 (252B/142B for 4B/8B).
We also observe consistent gains from 4B to 8B, highlighting the benefit of scaling under the block-wise discrete diffusion formulation.

\begin{table}[htbp]
\centering
\caption{\textbf{Comparison of reasoning model performance before and after long Chain-of-Thought (CoT) distillation.} `Inst' denotes models prior to long CoT distillation, while models with long CoT are indicated accordingly. 
R1-OneVision is a long-CoT-distilled variant built upon a Qwen2.5-VL-7B model.}
\label{tab:reasoning-model}
\setlength{\tabcolsep}{4pt}
\resizebox{\columnwidth}{!}{%
\begin{tabular}{l|ccc}
\toprule
\multirow{2}{*}{\textbf{Model}}& \textbf{MathVista} & \textbf{MathVerse} & \textbf{MathVision} \\
\cmidrule(l){2-4}
& mini & mini-vision & testmini \\
\midrule
Qwen2.5-VL-7B*      & 63.7 & 38.2 & 25.4  \\
R1-Onevision-7B*    & 64.1 ($\uparrow$0.6\%) & 40.0 ($\uparrow$4.7\%) & \textbf{29.9 ($\uparrow$17.7\%)} \\
\midrule
Lavida-8B*          & 44.8 & 27.2 & 20.4 \\
Lavida-8B-Reason*   & 45.2 ($\uparrow$0.9\%) & 29.3 ($\uparrow$7.7\%) & 24.0 ($\uparrow$17.6\%) \\
\midrule
SDAR-VL-4B-Inst     & 57.6 & 33.5 & 21.3 \\
SDAR-VL-4B-Think    & 60.4 ($\uparrow$4.9\%) & 38.6 ($\uparrow$15.2\%) & 25.7 ($\uparrow$20.7\%) \\
SDAR-VL-8B-Inst     & 62.5 & 36.5 & 24.2 \\
SDAR-VL-8B-Think    & \textbf{64.2 ($\uparrow$2.7\%)} & \textbf{41.8 ($\uparrow$14.5\%)} & 28.0 ($\uparrow$15.7\%) \\
\bottomrule
\end{tabular}
}
\end{table}
\paragraph{Multi-image and video benchmarks.}
Table~\ref{tab:multi-video-bench} further shows that SDAR-VL generalizes effectively to multi-image and video understanding. 
Within the controlled group, our 8B model achieves the best performance on BLINK, VideoMME, and LVBench, and remains competitive on MLVU, demonstrating strong cross-image and cross-frame aggregation ability. 
Compared with the matched diffusion baseline LLaDA-V, SDAR-VL shows consistent improvements across shared tasks, suggesting that block-wise diffusion provides a stronger foundation for multi-context visual reasoning.
Against broader open-source baselines, our results are competitive overall, outperforming Qwen2.5-VL on LVBench and MLVU while remaining close on VideoMME.

\paragraph{Effect of long CoT distillation.}
Table~\ref{tab:reasoning-model} shows that SDAR-VL consistently benefits from long CoT distillation, with clear gains on MathVista, MathVerse, and MathVision for both 4B and 8B variants. 
After distillation, SDAR-VL reaches strong competitiveness against CoT-enhanced autoregressive baselines, achieving near-parity on MathVision and surpassing R1-OneVision on MathVista/MathVerse. 
In contrast, the full discrete diffusion baseline (LaViDA) exhibits smaller improvements, suggesting long CoT signals are easier to exploit under the block-wise formulation. 
Overall, these results indicate that SDAR-VL offers a more effective interface to distill long-form reasoning into diffusion-based multimodal models.

\begin{figure}[htbp]
    \centering
    \includegraphics[width=0.48\textwidth]{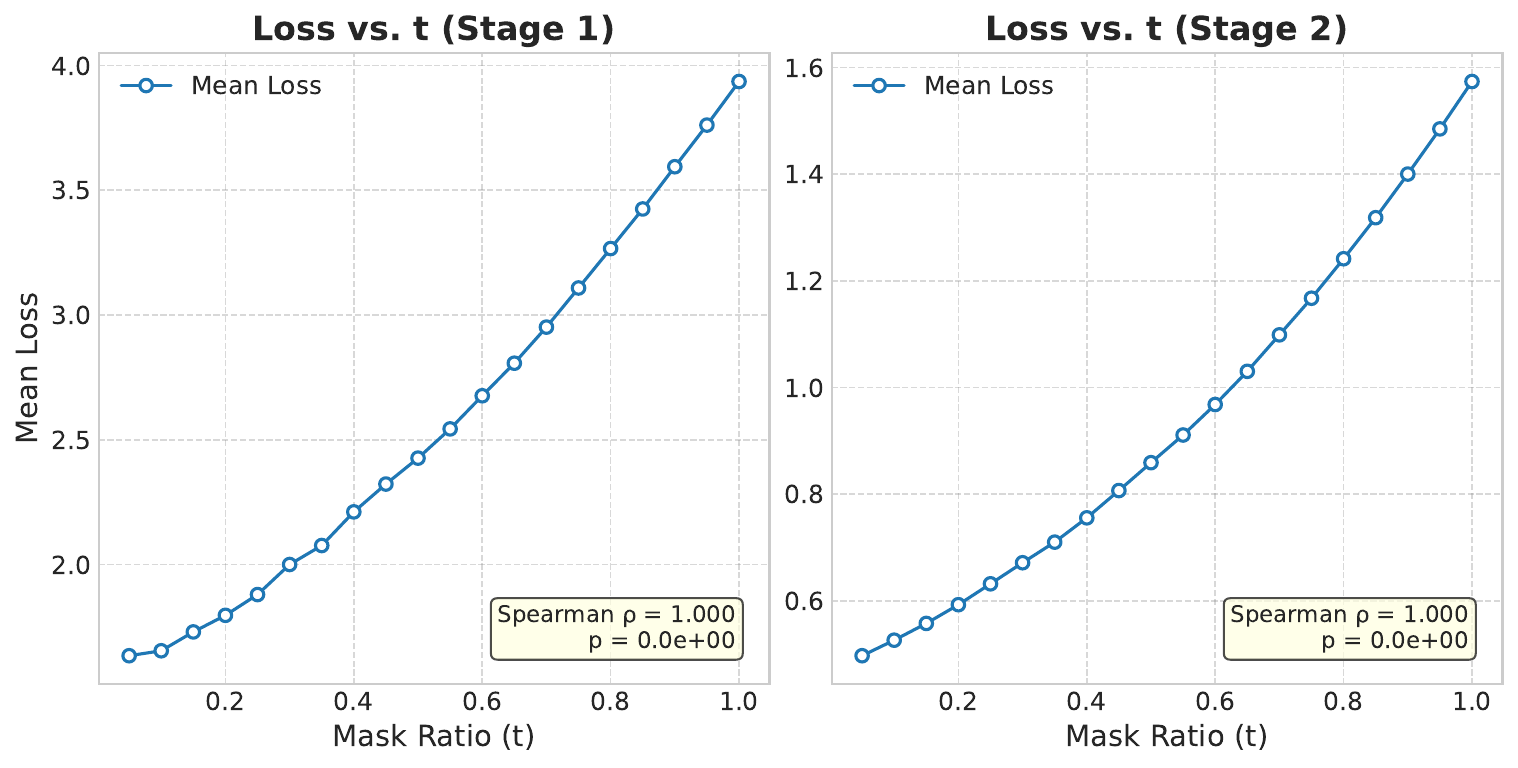}
    \caption{\textbf{Correlation between mask ratio \(t\) and average training loss at two stages.} Each point is the mean loss over samples with a fixed \(t\); loss grows with \(t\), indicating that higher noise increases reconstruction difficulty and motivating asynchronous noise scheduling for more stable training.}
    \label{fig:relation_t_loss}
\end{figure}

\begin{figure}[htbp]
    \centering
    \includegraphics[width=0.48\textwidth]{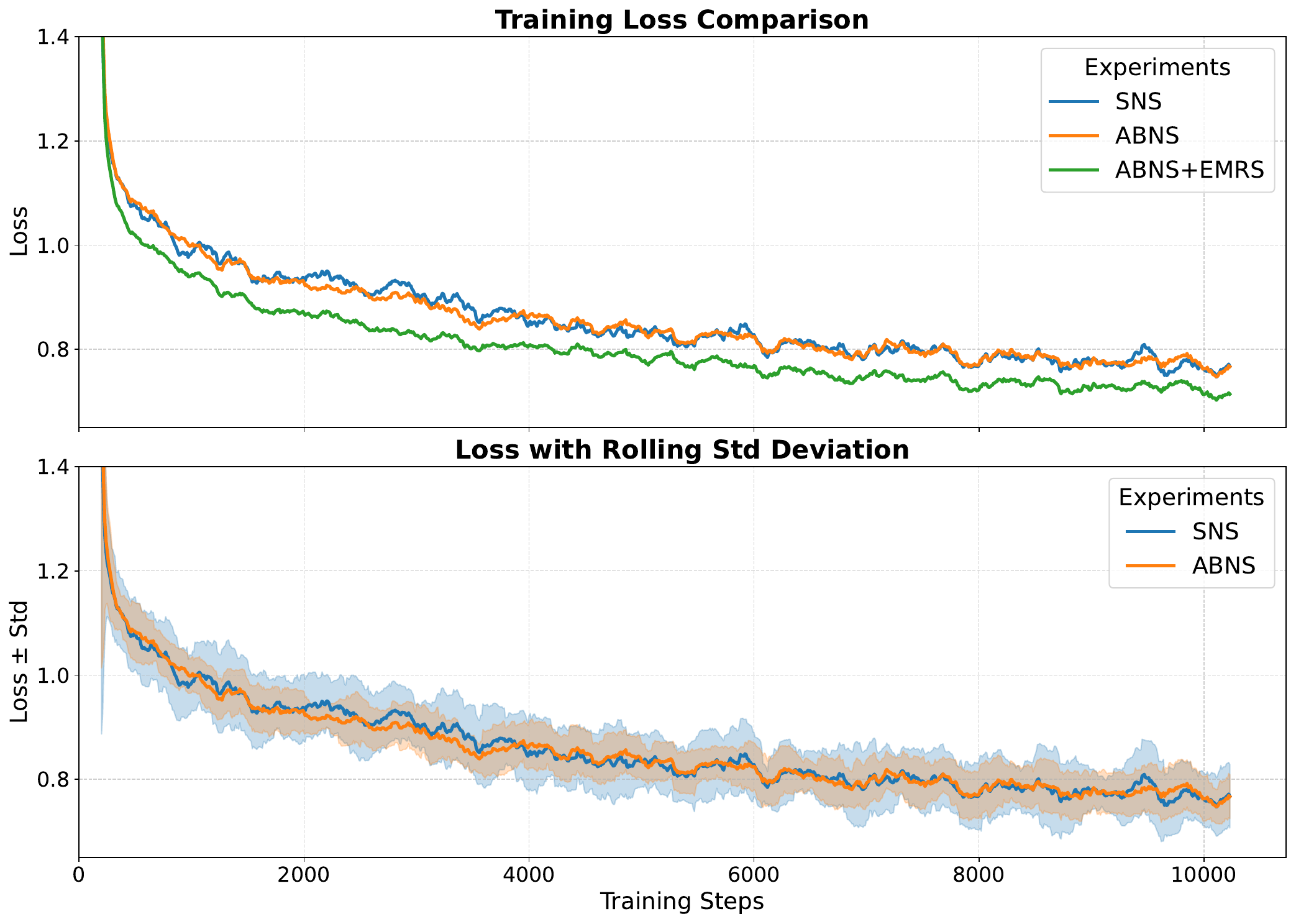}
    \caption{\textbf{Training loss of ablation variants on the 4B model during LLaVA-NeXT fine-tuning.} Top: training loss for synchronous noise scheduling (SNS), asynchronous scheduling (ABNS), and ABNS with effective mask ratio scaling (EMRS). Bottom: loss variance for SNS and ABNS, showing reduced fluctuation and improved stability with ABNS.}
    \label{fig:ablation_loss_curves}
\end{figure}
\subsection{Ablation Studies}
\label{sec:ablations}

To isolate the individual and combined effects of the proposed SDAR-VL components (Sec.~\ref{sec:method}), we perform controlled ablations on the 4B variant. 
All models follow the same two-stage protocol (Sec.~\ref{Implementation_Details}): projector-only alignment and subsequent full-parameter fine-tuning on LLaVA-NeXT.

\begin{table*}[htbp]
\centering
\caption{\textbf{Ablation study of SDAR-VL components on the 4B model.} We progressively add each component and evaluate performance on nine benchmarks after fine-tuning on LLaVA-NeXT. Scores are reported as percentages. The best result in each column is \textbf{bolded}. All models in this table use the same training budget.}
\label{tab:ablation_4b}
\resizebox{\textwidth}{!}{
\begin{tabular}{@{}l|ccccccccc@{}}
\toprule
\multirow{2}{*}{\textbf{Method}} & \textbf{SEEDBench} & \textbf{MMStar} & \textbf{MME} & \textbf{DocVQA} & \textbf{ChartQA} & \textbf{AI2D} & \textbf{MathVista} & \textbf{ScienceQA} & \textbf{HallBench} \\
    \cmidrule(l){2-10}
    & {image} & {test} & {perp/cog} & {val} & {test} & {test} & {mini} & {test} & {avg} \\ 
\midrule
Baseline (SNS)                    & 71.6 & 46.2 & 1480/326 & 68.2 & 65.4 & 71.9 & 41.6 & 75.4 & 36.8  \\
ABNS                              & 71.8 & 47.0 & 1482/327 & 68.1 & 67.4 & 72.0 & 40.2 & 74.9 & 37.5  \\
ABNS + EMRS                       & 72.1 & 47.5 & 1489/344 & 68.8 & 68.9 & 74.6 & 42.2 & 77.6 & 37.8  \\
ABNS + EMRS + Clamp               & 72.6 & 48.2 & 1490/344 & \textbf{70.2} & 68.1 & 73.6 & 41.9 & 75.4 & 36.5  \\
\midrule
\textbf{ABNS + EMRS + PBNC (c=25)} & \textbf{73.5} & 47.7 & 1501/331 & 69.4 & \textbf{69.3} & 74.5 & \textbf{43.8} & \textbf{77.8} & 38.4  \\
\textbf{ABNS + EMRS + PBNC (c=50)} & \textbf{73.5} & \textbf{48.5} & \textbf{1515/336} & \textbf{70.2} & 68.2 & \textbf{74.4} & 42.7 & 77.6 & \textbf{40.1}  \\
\bottomrule
\end{tabular}
}
\end{table*}

We start from a BD3~\cite{arriola2025block,cheng2025sdar} baseline with \emph{synchronous noise scheduling} (SNS), where a single mask ratio $t$ is sampled per step for all blocks with $1/t$ loss scaling. 
We then introduce a sequence of incremental variants. 
\emph{ABNS} replaces SNS with Asynchronous Block-wise Noise Scheduling (Sec.~\ref{sec:asynchronous-noise-scheduler}), independently sampling a noise ratio $t_b$ for each block to increase in-step supervision diversity. 
\emph{ABNS+EMRS} further applies Effective Mask Ratio Scaling (Sec.~\ref{sec:effective-mask-scaler}), normalizing the loss by the realized mask ratio $t'_b$ instead of the sampled $t_b$ to reduce bias and gradient variance. 
\emph{ABNS+EMRS+Clamp} serves as a strong static baseline that restricts the noise distribution to a high-noise uniform range $t_b \sim \mathcal{U}(0.45, 0.95)$, emphasizing challenging reconstructions but reducing exposure to low-noise cases. 
Finally, \emph{ABNS+EMRS+PBNC} corresponds to the full SDAR-VL configuration and adopts the Progressive Beta Distribution Noise Curriculum (PBNC, Sec.~\ref{sec:progressive-beta}), which gradually increases both the mean and concentration $(\mu_\tau, C_\tau)$ of the noise distribution to balance diversity and difficulty; we report results for two curriculum sharpness settings with $C_{\text{final}}{=}25$ and $C_{\text{final}}{=}50$.

\subsubsection{Mask Ratio-Loss Analysis}

To motivate asynchronous noise scheduling, we analyze how reconstruction difficulty varies with the mask ratio \(t\). 
Using checkpoints from two stages (the initial stage and LLaVA-NeXT fine-tuning), we evaluate a fixed 10k subset of LLaVA-NeXT and compute the mean loss for each discrete \(t\). 

As shown in Figure~\ref{fig:relation_t_loss}, the loss increases monotonically with \(t\), with strong positive Spearman correlations in both stages. 
This indicates that \textbf{higher mask ratios correspond to harder reconstructions and thus larger losses.} 
Consequently, under synchronous scheduling, sampling a single \(t\) per step can translate directly into substantial step-to-step loss variance, potentially destabilizing optimization. 
In contrast, asynchronous block-wise noise scheduling (ABNS) assigns diverse mask ratios \(t_b\) across blocks within a step, effectively averaging over a richer \(t\) distribution and smoothing loss fluctuations to improve training stability.

\subsubsection{Training Loss Dynamics}

We next analyze training loss curves of three configurations: synchronous scheduler (SNS), asynchronous block-wise scheduler (ABNS), and ABNS with mask ratio scaling (EMRS).

Figure~\ref{fig:ablation_loss_curves} (top) shows the average training loss during LLaVA-NeXT fine-tuning. 
ABNS and SNS exhibit similar mean loss trends, whereas EMRS consistently lowers the loss and improves convergence stability, indicating more effective optimization via unbiased loss scaling. 
The bottom subplot shows that ABNS markedly reduces loss variance despite comparable mean loss, underscoring its key role in stabilizing training.

Due to the strong dependence of loss magnitude on \(t\), plotting loss curves across different noise schedulers with varying mask ratio scalings (e.g., clamp or progressive beta) is not meaningful. Hence, those are excluded from loss curve plots but evaluated in downstream benchmarks.

\subsubsection{Downstream Task Performance}

Table~\ref{tab:ablation_4b} presents ablations on the 4B model with identical training budget across nine benchmarks. Relative to SNS, \textbf{ABNS} yields modest gains on ChartQA, MMStar, and HallBench, indicating limited but positive effects from increased in-step noise diversity. \textbf{EMRS} brings broader improvements across most tasks, notably on MME cognition and ScienceQA, suggesting more stable optimization via unbiased scaling. \textbf{Clamp} improves select scores (e.g., DocVQA) but hurts robustness, revealing a non-trivial coverage--diversity trade-off. \textbf{PBNC} best resolves this trade-off and delivers the strongest overall performance; $C_{\text{final}}{=}25$ tends to favor reasoning- and text-rich tasks, while $C_{\text{final}}{=}50$ strengthens general understanding and hallucination robustness. Overall, ABNS, EMRS, and PBNC provide complementary benefits, and the full SDAR-VL setup achieves the best results.

\section{Conclusion}
\label{sec:conclusion}

We introduced SDAR-VL, a block-wise discrete diffusion framework for vision-language understanding that stabilizes BD3 training via asynchronous block-wise noise scheduling, effective mask ratio scaling, and a progressive Beta noise curriculum. Across 21 benchmarks, SDAR-VL sets a new state of the art among diffusion-based vision-language models and, under matched settings, outperforms the diffusion baseline LLaDA-V as well as autoregressive LLaVA-OneVision. With long CoT distillation, SDAR-VL-Think further reaches parity with or surpasses CoT-enhanced autoregressive baselines such as the Qwen2.5-VL–based R1-OneVision on math-centric benchmarks, highlighting block-wise discrete diffusion as a practical backbone for multimodal reasoning.

\section{Acknowledge}
\label{sec:acknowledge}

Supported by Shanghai Artificial Intelligence Laboratory. In addition, this work was supported by the National Natural Science Foundation of China (Grant No.~6250076080) and the China Postdoctoral Science Foundation (Grant No.~2025M771537).

\bibliography{anthology}

@article{xin2025lumina,
  title={Lumina-dimoo: An omni diffusion large language model for multi-modal generation and understanding},
  author={Xin, Yi and Qin, Qi and Luo, Siqi and Zhu, Kaiwen and Yan, Juncheng and Tai, Yan and Lei, Jiayi and Cao, Yuewen and Wang, Keqi and Wang, Yibin and others},
  journal={arXiv preprint arXiv:2510.06308},
  year={2025}
}

@article{yang2025mmada,
  title={Mmada: Multimodal large diffusion language models},
  author={Yang, Ling and Tian, Ye and Li, Bowen and Zhang, Xinchen and Shen, Ke and Tong, Yunhai and Wang, Mengdi},
  journal={arXiv preprint arXiv:2505.15809},
  year={2025}
}

@article{you2025llada,
  title={Llada-v: Large language diffusion models with visual instruction tuning},
  author={You, Zebin and Nie, Shen and Zhang, Xiaolu and Hu, Jun and Zhou, Jun and Lu, Zhiwu and Wen, Ji-Rong and Li, Chongxuan},
  journal={arXiv preprint arXiv:2505.16933},
  year={2025}
}

@article{cheng2025sdar,
  title={SDAR: A Synergistic Diffusion-AutoRegression Paradigm for Scalable Sequence Generation},
  author={Cheng, Shuang and Bian, Yihan and Liu, Dawei and Jiang, Yuhua and Liu, Yihao and Zhang, Linfeng and Wang, Wenhai and Guo, Qipeng and Chen, Kai and Qi, Biqing and others},
  journal={arXiv preprint arXiv:2510.06303},
  year={2025}
}

@article{tschannen2025siglip,
  title={Siglip 2: Multilingual vision-language encoders with improved semantic understanding, localization, and dense features},
  author={Tschannen, Michael and Gritsenko, Alexey and Wang, Xiao and Naeem, Muhammad Ferjad and Alabdulmohsin, Ibrahim and Parthasarathy, Nikhil and Evans, Talfan and Beyer, Lucas and Xia, Ye and Mustafa, Basil and others},
  journal={arXiv preprint arXiv:2502.14786},
  year={2025}
}

@article{liu2023visual,
  title={Visual instruction tuning},
  author={Liu, Haotian and Li, Chunyuan and Wu, Qingyang and Lee, Yong Jae},
  journal={Advances in neural information processing systems},
  volume={36},
  pages={34892--34916},
  year={2023}
}

@article{guo2024mammoth,
  title={Mammoth-vl: Eliciting multimodal reasoning with instruction tuning at scale},
  author={Guo, Jarvis and Zheng, Tuney and Bai, Yuelin and Li, Bo and Wang, Yubo and Zhu, King and Li, Yizhi and Neubig, Graham and Chen, Wenhu and Yue, Xiang},
  journal={arXiv preprint arXiv:2412.05237},
  year={2024}
}

@article{jia2025visualwebinstruct,
  title={Visualwebinstruct: Scaling up multimodal instruction data through web search},
  author={Jia, Yiming and Li, Jiachen and Yue, Xiang and Li, Bo and Nie, Ping and Zou, Kai and Chen, Wenhu},
  journal={arXiv preprint arXiv:2503.10582},
  year={2025}
}

@misc{liu2024llavanext,
  title={Llavanext: Improved reasoning, ocr, and world knowledge},
  author={Liu, Haotian and Li, Chunyuan and Li, Yuheng and Li, Bo and Zhang, Yuanhan and Shen, Sheng and Lee, Yong Jae},
  year={2024}
}

@article{arriola2025block,
  title={Block diffusion: Interpolating between autoregressive and diffusion language models},
  author={Arriola, Marianne and Gokaslan, Aaron and Chiu, Justin T and Yang, Zhihan and Qi, Zhixuan and Han, Jiaqi and Sahoo, Subham Sekhar and Kuleshov, Volodymyr},
  journal={arXiv preprint arXiv:2503.09573},
  year={2025}
}

@article{zhu2025lladamoe,
  title={LLaDA-MoE: A Sparse MoE Diffusion Language Model},
  author={Zhu, Fengqi and You, Zebin and Xing, Yipeng and Huang, Zenan and Liu, Lin and Zhuang, Yihong and Lu, Guoshan and Wang, Kangyu and Wang, Xudong and Wei, Lanning and others},
  journal={arXiv preprint arXiv:2509.24389},
  year={2025}
}

@inproceedings{lou2024discrete,
  title={Discrete diffusion modeling by estimating the ratios of the data distribution},
  author={Lou, Aaron and Meng, Chenlin and Ermon, Stefano},
  booktitle={Proceedings of the 41st International Conference on Machine Learning},
  pages={32819--32848},
  year={2024}
}

@article{austin2021structured,
  title={Structured denoising diffusion models in discrete state-spaces},
  author={Austin, Jacob and Johnson, Daniel D and Ho, Jonathan and Tarlow, Daniel and Van Den Berg, Rianne},
  journal={Advances in neural information processing systems},
  volume={34},
  pages={17981--17993},
  year={2021}
}

@article{gulrajani2023likelihood,
  title={Likelihood-based diffusion language models},
  author={Gulrajani, Ishaan and Hashimoto, Tatsunori B},
  journal={Advances in Neural Information Processing Systems},
  volume={36},
  pages={16693--16715},
  year={2023}
}

@article{sahoo2024simple,
  title={Simple and effective masked diffusion language models},
  author={Sahoo, Subham and Arriola, Marianne and Schiff, Yair and Gokaslan, Aaron and Marroquin, Edgar and Chiu, Justin and Rush, Alexander and Kuleshov, Volodymyr},
  journal={Advances in Neural Information Processing Systems},
  volume={37},
  pages={130136--130184},
  year={2024}
}

@article{nie2025large,
  title={Large language diffusion models},
  author={Nie, Shen and Zhu, Fengqi and You, Zebin and Zhang, Xiaolu and Ou, Jingyang and Hu, Jun and Zhou, Jun and Lin, Yankai and Wen, Ji-Rong and Li, Chongxuan},
  journal={arXiv preprint arXiv:2502.09992},
  year={2025}
}

@article{gong2024scaling,
  title={Scaling diffusion language models via adaptation from autoregressive models},
  author={Gong, Shansan and Agarwal, Shivam and Zhang, Yizhe and Ye, Jiacheng and Zheng, Lin and Li, Mukai and An, Chenxin and Zhao, Peilin and Bi, Wei and Han, Jiawei and others},
  journal={arXiv preprint arXiv:2410.17891},
  year={2024}
}

@article{zhu2025llada,
  title={LLaDA 1.5: Variance-Reduced Preference Optimization for Large Language Diffusion Models},
  author={Zhu, Fengqi and Wang, Rongzhen and Nie, Shen and Zhang, Xiaolu and Wu, Chunwei and Hu, Jun and Zhou, Jun and Chen, Jianfei and Lin, Yankai and Wen, Ji-Rong and others},
  journal={arXiv preprint arXiv:2505.19223},
  year={2025}
}

@article{ye2025dream,
  title={Dream 7b: Diffusion large language models},
  author={Ye, Jiacheng and Xie, Zhihui and Zheng, Lin and Gao, Jiahui and Wu, Zirui and Jiang, Xin and Li, Zhenguo and Kong, Lingpeng},
  journal={arXiv preprint arXiv:2508.15487},
  year={2025}
}

@article{ye2024beyond,
  title={Beyond autoregression: Discrete diffusion for complex reasoning and planning},
  author={Ye, Jiacheng and Gao, Jiahui and Gong, Shansan and Zheng, Lin and Jiang, Xin and Li, Zhenguo and Kong, Lingpeng},
  journal={arXiv preprint arXiv:2410.14157},
  year={2024}
}

@article{wang2025diffusion,
  title={Diffusion llms can do faster-than-ar inference via discrete diffusion forcing},
  author={Wang, Xu and Xu, Chenkai and Jin, Yijie and Jin, Jiachun and Zhang, Hao and Deng, Zhijie},
  journal={arXiv preprint arXiv:2508.09192},
  year={2025}
}

@article{wu2025fastv2,
  title={Fast-dllm v2: Efficient block-diffusion llm},
  author={Wu, Chengyue and Zhang, Hao and Xue, Shuchen and Diao, Shizhe and Fu, Yonggan and Liu, Zhijian and Molchanov, Pavlo and Luo, Ping and Han, Song and Xie, Enze},
  journal={arXiv preprint arXiv:2509.26328},
  year={2025}
}

@article{zhang2025survey,
  title={A survey on parallel text generation: From parallel decoding to diffusion language models},
  author={Zhang, Lingzhe and Fang, Liancheng and Duan, Chiming and He, Minghua and Pan, Leyi and Xiao, Pei and Huang, Shiyu and Zhai, Yunpeng and Hu, Xuming and Yu, Philip S and others},
  journal={arXiv preprint arXiv:2508.08712},
  year={2025}
}

@article{yu2025discrete,
  title={Discrete Diffusion in Large Language and Multimodal Models: A Survey},
  author={Yu, Runpeng and Li, Qi and Wang, Xinchao},
  journal={arXiv preprint arXiv:2506.13759},
  year={2025}
}

@article{ji2025denoising,
  title={From Denoising to Refining: A Corrective Framework for Vision-Language Diffusion Model},
  author={Ji, Yatai and Wang, Teng and Ge, Yuying and Liu, Zhiheng and Yang, Sidi and Shan, Ying and Luo, Ping},
  journal={arXiv preprint arXiv:2510.19871},
  year={2025}
}

@article{ye2024diffusion,
  title={Diffusion of thought: Chain-of-thought reasoning in diffusion language models},
  author={Ye, Jiacheng and Gong, Shansan and Chen, Liheng and Zheng, Lin and Gao, Jiahui and Shi, Han and Wu, Chuan and Jiang, Xin and Li, Zhenguo and Bi, Wei and others},
  journal={Advances in Neural Information Processing Systems},
  volume={37},
  pages={105345--105374},
  year={2024}
}

@article{li2025lavida,
  title={Lavida: A large diffusion language model for multimodal understanding},
  author={Li, Shufan and Kallidromitis, Konstantinos and Bansal, Hritik and Gokul, Akash and Kato, Yusuke and Kozuka, Kazuki and Kuen, Jason and Lin, Zhe and Chang, Kai-Wei and Grover, Aditya},
  journal={arXiv preprint arXiv:2505.16839},
  year={2025}
}

@article{yu2025dimple,
  title={Dimple: Discrete diffusion multimodal large language model with parallel decoding},
  author={Yu, Runpeng and Ma, Xinyin and Wang, Xinchao},
  journal={arXiv preprint arXiv:2505.16990},
  year={2025}
}

@article{li2025lavidao,
  title={Lavida-O: Elastic Large Masked Diffusion Models for Unified Multimodal Understanding and Generation},
  author={Li, Shufan and Gu, Jiuxiang and Liu, Kangning and Lin, Zhe and Wei, Zijun and Grover, Aditya and Kuen, Jason},
  journal={arXiv e-prints},
  pages={arXiv--2509},
  year={2025}
}

@article{shi2025muddit,
  title={Muddit: Liberating generation beyond text-to-image with a unified discrete diffusion model},
  author={Shi, Qingyu and Bai, Jinbin and Zhao, Zhuoran and Chai, Wenhao and Yu, Kaidong and Wu, Jianzong and Song, Shuangyong and Tong, Yunhai and Li, Xiangtai and Li, Xuelong and others},
  journal={arXiv preprint arXiv:2505.23606},
  year={2025}
}

@inproceedings{liu2024mmbench,
  title={Mmbench: Is your multi-modal model an all-around player?},
  author={Liu, Yuan and Duan, Haodong and Zhang, Yuanhan and Li, Bo and Zhang, Songyang and Zhao, Wangbo and Yuan, Yike and Wang, Jiaqi and He, Conghui and Liu, Ziwei and others},
  booktitle={European conference on computer vision},
  pages={216--233},
  year={2024},
  organization={Springer}
}

@misc{fu2025mmecomprehensiveevaluationbenchmark,
      title={MME: A Comprehensive Evaluation Benchmark for Multimodal Large Language Models}, 
      author={Chaoyou Fu and Peixian Chen and Yunhang Shen and Yulei Qin and Mengdan Zhang and Xu Lin and Jinrui Yang and Xiawu Zheng and Ke Li and Xing Sun and Yunsheng Wu and Rongrong Ji and Caifeng Shan and Ran He},
      year={2025},
      eprint={2306.13394},
      archivePrefix={arXiv},
      primaryClass={cs.CV},
      url={https://arxiv.org/abs/2306.13394}, 
}

@article{chen2024we,
  title={Are we on the right way for evaluating large vision-language models?},
  author={Chen, Lin and Li, Jinsong and Dong, Xiaoyi and Zhang, Pan and Zang, Yuhang and Chen, Zehui and Duan, Haodong and Wang, Jiaqi and Qiao, Yu and Lin, Dahua and others},
  journal={Advances in Neural Information Processing Systems},
  volume={37},
  pages={27056--27087},
  year={2024}
}

@inproceedings{yue2024mmmu,
  title={Mmmu: A massive multi-discipline multimodal understanding and reasoning benchmark for expert agi},
  author={Yue, Xiang and Ni, Yuansheng and Zhang, Kai and Zheng, Tianyu and Liu, Ruoqi and Zhang, Ge and Stevens, Samuel and Jiang, Dongfu and Ren, Weiming and Sun, Yuxuan and others},
  booktitle={Proceedings of the IEEE/CVF Conference on Computer Vision and Pattern Recognition},
  pages={9556--9567},
  year={2024}
}

@article{lu2023mathvista,
  title={Mathvista: Evaluating math reasoning in visual contexts with gpt-4v, bard, and other large multimodal models},
  author={Lu, Pan and Bansal, Hritik and Xia, Tony and Liu, Jiacheng and Li, Chunyuan and Hajishirzi, Hannaneh and Cheng, Hao and Chang, Kai-Wei and Galley, Michel and Gao, Jianfeng},
  journal={CoRR},
  year={2023}
}

@inproceedings{zhang2024mathverse,
  title={Mathverse: Does your multi-modal llm truly see the diagrams in visual math problems?},
  author={Zhang, Renrui and Jiang, Dongzhi and Zhang, Yichi and Lin, Haokun and Guo, Ziyu and Qiu, Pengshuo and Zhou, Aojun and Lu, Pan and Chang, Kai-Wei and Qiao, Yu and others},
  booktitle={European Conference on Computer Vision},
  pages={169--186},
  year={2024},
  organization={Springer}
}

@inproceedings{mathew2021docvqa,
  title={Docvqa: A dataset for vqa on document images},
  author={Mathew, Minesh and Karatzas, Dimosthenis and Jawahar, CV},
  booktitle={Proceedings of the IEEE/CVF winter conference on applications of computer vision},
  pages={2200--2209},
  year={2021}
}

@article{masry2022chartqa,
  title={Chartqa: A benchmark for question answering about charts with visual and logical reasoning},
  author={Masry, Ahmed and Long, Do Xuan and Tan, Jia Qing and Joty, Shafiq and Hoque, Enamul},
  journal={arXiv preprint arXiv:2203.10244},
  year={2022}
}

@inproceedings{kembhavi2016diagram,
  title={A diagram is worth a dozen images},
  author={Kembhavi, Aniruddha and Salvato, Mike and Kolve, Eric and Seo, Minjoon and Hajishirzi, Hannaneh and Farhadi, Ali},
  booktitle={European conference on computer vision},
  pages={235--251},
  year={2016},
  organization={Springer}
}

@inproceedings{mathew2022infographicvqa,
  title={Infographicvqa},
  author={Mathew, Minesh and Bagal, Viraj and Tito, Rub{\`e}n and Karatzas, Dimosthenis and Valveny, Ernest and Jawahar, CV},
  booktitle={Proceedings of the IEEE/CVF Winter Conference on Applications of Computer Vision},
  pages={1697--1706},
  year={2022}
}

@article{wang2024muirbench,
  title={Muirbench: A comprehensive benchmark for robust multi-image understanding},
  author={Wang, Fei and Fu, Xingyu and Huang, James Y and Li, Zekun and Liu, Qin and Liu, Xiaogeng and Ma, Mingyu Derek and Xu, Nan and Zhou, Wenxuan and Zhang, Kai and others},
  journal={arXiv preprint arXiv:2406.09411},
  year={2024}
}

@article{zhou2024mlvu,
  title={Mlvu: A comprehensive benchmark for multi-task long video understanding},
  author={Zhou, Junjie and Shu, Yan and Zhao, Bo and Wu, Boya and Xiao, Shitao and Yang, Xi and Xiong, Yongping and Zhang, Bo and Huang, Tiejun and Liu, Zheng},
  journal={arXiv e-prints},
  pages={arXiv--2406},
  year={2024}
}

@inproceedings{fu2025video,
  title={Video-mme: The first-ever comprehensive evaluation benchmark of multi-modal llms in video analysis},
  author={Fu, Chaoyou and Dai, Yuhan and Luo, Yongdong and Li, Lei and Ren, Shuhuai and Zhang, Renrui and Wang, Zihan and Zhou, Chenyu and Shen, Yunhang and Zhang, Mengdan and others},
  booktitle={Proceedings of the Computer Vision and Pattern Recognition Conference},
  pages={24108--24118},
  year={2025}
}

@article{li2023seed,
  title={Seed-bench: Benchmarking multimodal llms with generative comprehension},
  author={Li, Bohao and Wang, Rui and Wang, Guangzhi and Ge, Yuying and Ge, Yixiao and Shan, Ying},
  journal={arXiv preprint arXiv:2307.16125},
  year={2023}
}

@article{yu2023mm,
  title={Mm-vet: Evaluating large multimodal models for integrated capabilities},
  author={Yu, Weihao and Yang, Zhengyuan and Li, Linjie and Wang, Jianfeng and Lin, Kevin and Liu, Zicheng and Wang, Xinchao and Wang, Lijuan},
  journal={arXiv preprint arXiv:2308.02490},
  year={2023}
}

@article{peng2025efficient,
  title={How Efficient Are Diffusion Language Models? A Critical Examination of Efficiency Evaluation Practices},
  author={Peng, Han and Liu, Peiyu and Dong, Zican and Cheng, Daixuan and Li, Junyi and Tang, Yiru and Wang, Shuo and Zhao, Wayne Xin},
  journal={arXiv preprint arXiv:2510.18480},
  year={2025}
}

@article{lu2022learn,
  title={Learn to explain: Multimodal reasoning via thought chains for science question answering},
  author={Lu, Pan and Mishra, Swaroop and Xia, Tanglin and Qiu, Liang and Chang, Kai-Wei and Zhu, Song-Chun and Tafjord, Oyvind and Clark, Peter and Kalyan, Ashwin},
  journal={Advances in Neural Information Processing Systems},
  volume={35},
  pages={2507--2521},
  year={2022}
}

@inproceedings{fu2024blink,
  title={Blink: Multimodal large language models can see but not perceive},
  author={Fu, Xingyu and Hu, Yushi and Li, Bangzheng and Feng, Yu and Wang, Haoyu and Lin, Xudong and Roth, Dan and Smith, Noah A and Ma, Wei-Chiu and Krishna, Ranjay},
  booktitle={European Conference on Computer Vision},
  pages={148--166},
  year={2024},
  organization={Springer}
}

@inproceedings{li2024mvbench,
  title={Mvbench: A comprehensive multi-modal video understanding benchmark},
  author={Li, Kunchang and Wang, Yali and He, Yinan and Li, Yizhuo and Wang, Yi and Liu, Yi and Wang, Zun and Xu, Jilan and Chen, Guo and Luo, Ping and others},
  booktitle={Proceedings of the IEEE/CVF Conference on Computer Vision and Pattern Recognition},
  pages={22195--22206},
  year={2024}
}

@inproceedings{guan2024hallusionbench,
  title={Hallusionbench: an advanced diagnostic suite for entangled language hallucination and visual illusion in large vision-language models},
  author={Guan, Tianrui and Liu, Fuxiao and Wu, Xiyang and Xian, Ruiqi and Li, Zongxia and Liu, Xiaoyu and Wang, Xijun and Chen, Lichang and Huang, Furong and Yacoob, Yaser and others},
  booktitle={Proceedings of the IEEE/CVF Conference on Computer Vision and Pattern Recognition},
  pages={14375--14385},
  year={2024}
}

@article{wang2024measuring,
  title={Measuring multimodal mathematical reasoning with math-vision dataset},
  author={Wang, Ke and Pan, Junting and Shi, Weikang and Lu, Zimu and Ren, Houxing and Zhou, Aojun and Zhan, Mingjie and Li, Hongsheng},
  journal={Advances in Neural Information Processing Systems},
  volume={37},
  pages={95095--95169},
  year={2024}
}

@article{yang2025r1,
  title={R1-onevision: Advancing generalized multimodal reasoning through cross-modal formalization},
  author={Yang, Yi and He, Xiaoxuan and Pan, Hongkun and Jiang, Xiyan and Deng, Yan and Yang, Xingtao and Lu, Haoyu and Yin, Dacheng and Rao, Fengyun and Zhu, Minfeng and others},
  journal={arXiv preprint arXiv:2503.10615},
  year={2025}
}

@misc{opengvlab2024internvl2,
  title={InternVL2: Better than the Best—Expanding Performance Boundaries of Open-Source Multimodal Models with the Progressive Scaling Strategy},
  author={{OpenGVLab Team}},
  year={2024},
  month={July},
  day={4},
  note={Accessed 2024-04-27},
  howpublished={\url{https://internvl.github.io/blog/2024-07-02-InternVL-2.0/}}
}

@article{li2024llava,
  title={Llava-onevision: Easy visual task transfer},
  author={Li, Bo and Zhang, Yuanhan and Guo, Dong and Zhang, Renrui and Li, Feng and Zhang, Hao and Zhang, Kaichen and Zhang, Peiyuan and Li, Yanwei and Liu, Ziwei and others},
  journal={arXiv preprint arXiv:2408.03326},
  year={2024}
}

@article{wang2024qwen2,
  title={Qwen2-vl: Enhancing vision-language model's perception of the world at any resolution},
  author={Wang, Peng and Bai, Shuai and Tan, Sinan and Wang, Shijie and Fan, Zhihao and Bai, Jinze and Chen, Keqin and Liu, Xuejing and Wang, Jialin and Ge, Wenbin and others},
  journal={arXiv preprint arXiv:2409.12191},
  year={2024}
}

@article{bai2025qwen2,
  title={Qwen2. 5-vl technical report},
  author={Bai, Shuai and Chen, Keqin and Liu, Xuejing and Wang, Jialin and Ge, Wenbin and Song, Sibo and Dang, Kai and Wang, Peng and Wang, Shijie and Tang, Jun and others},
  journal={arXiv preprint arXiv:2502.13923},
  year={2025}
}

@article{chen2024expanding,
  title={Expanding performance boundaries of open-source multimodal models with model, data, and test-time scaling},
  author={Chen, Zhe and Wang, Weiyun and Cao, Yue and Liu, Yangzhou and Gao, Zhangwei and Cui, Erfei and Zhu, Jinguo and Ye, Shenglong and Tian, Hao and Liu, Zhaoyang and others},
  journal={arXiv preprint arXiv:2412.05271},
  year={2024}
}

@article{wang2025internvl3,
  title={Internvl3. 5: Advancing open-source multimodal models in versatility, reasoning, and efficiency},
  author={Wang, Weiyun and Gao, Zhangwei and Gu, Lixin and Pu, Hengjun and Cui, Long and Wei, Xingguang and Liu, Zhaoyang and Jing, Linglin and Ye, Shenglong and Shao, Jie and others},
  journal={arXiv preprint arXiv:2508.18265},
  year={2025}
}

@article{comanici2025gemini,
  title={Gemini 2.5: Pushing the frontier with advanced reasoning, multimodality, long context, and next generation agentic capabilities},
  author={Comanici, Gheorghe and Bieber, Eric and Schaekermann, Mike and Pasupat, Ice and Sachdeva, Noveen and Dhillon, Inderjit and Blistein, Marcel and Ram, Ori and Zhang, Dan and Rosen, Evan and others},
  journal={arXiv preprint arXiv:2507.06261},
  year={2025}
}

@article{hurst2024gpt,
  title={Gpt-4o system card},
  author={Hurst, Aaron and Lerer, Adam and Goucher, Adam P and Perelman, Adam and Ramesh, Aditya and Clark, Aidan and Ostrow, AJ and Welihinda, Akila and Hayes, Alan and Radford, Alec and others},
  journal={arXiv preprint arXiv:2410.21276},
  year={2024}
}

@article{wu2025fast,
  title={Fast-dllm: Training-free acceleration of diffusion llm by enabling kv cache and parallel decoding},
  author={Wu, Chengyue and Zhang, Hao and Xue, Shuchen and Liu, Zhijian and Diao, Shizhe and Zhu, Ligeng and Luo, Ping and Han, Song and Xie, Enze},
  journal={arXiv preprint arXiv:2505.22618},
  year={2025}
}

@article{ma2025dkv,
  title={dkv-cache: The cache for diffusion language models},
  author={Ma, Xinyin and Yu, Runpeng and Fang, Gongfan and Wang, Xinchao},
  journal={arXiv preprint arXiv:2505.15781},
  year={2025}
}

@article{yang2025diffusion,
  title={Diffusion LLM with Native Variable Generation Lengths: Let [EOS] Lead the Way},
  author={Yang, Yicun and Wang, Cong and Wang, Shaobo and Wen, Zichen and Qi, Biqing and Xu, Hanlin and Zhang, Linfeng},
  journal={arXiv preprint arXiv:2510.24605},
  year={2025}
}

@misc{duan2025vlmevalkitopensourcetoolkitevaluating,
      title={VLMEvalKit: An Open-Source Toolkit for Evaluating Large Multi-Modality Models}, 
      author={Haodong Duan and Xinyu Fang and Junming Yang and Xiangyu Zhao and Yuxuan Qiao and Mo Li and Amit Agarwal and Zhe Chen and Lin Chen and Yuan Liu and Yubo Ma and Hailong Sun and Yifan Zhang and Shiyin Lu and Tack Hwa Wong and Weiyun Wang and Peiheng Zhou and Xiaozhe Li and Chaoyou Fu and Junbo Cui and Jixuan Chen and Enxin Song and Song Mao and Shengyuan Ding and Tianhao Liang and Zicheng Zhang and Xiaoyi Dong and Yuhang Zang and Pan Zhang and Jiaqi Wang and Dahua Lin and Kai Chen},
      year={2025},
      eprint={2407.11691},
      archivePrefix={arXiv},
      primaryClass={cs.CV},
      url={https://arxiv.org/abs/2407.11691}, 
}

\appendix

\section{Related Work}
\label{sec:related_work}

\subsection{Large Language Diffusion Models}
The paradigm of sequence generation has recently seen the emergence of diffusion models as a compelling alternative to traditional autoregressive approaches~\citep{nie2025large,gong2024scaling,zhu2025lladamoe,cheng2025sdar}. 
Unlike AR models that predict sequences token-by-token, discrete diffusion models treat generation as a denoising process, learning to reverse a gradual corruption of data~\cite{austin2021structured,gulrajani2023likelihood,lou2024discrete,sahoo2024simple,arriola2025block}. 
Among these, masked diffusion has become the \textit{de facto} standard, where the model learns to restore original tokens from a sequence corrupted with `[MASK]' tokens at various noise levels~\cite{austin2021structured, sahoo2024simple}.
This architectural shift offers several fundamental advantages. 
First, it enables parallel decoding, directly addressing the high latency inherent in the sequential nature of AR inference~\cite{zhang2025survey}. 
Second, by relaxing the strict left-to-right causal bias, diffusion models possess a bidirectional context, an inductive bias better suited for tasks requiring holistic reasoning and non-local dependency modeling~\cite{ye2024beyond,ye2025dream}. 
This flexibility also hints at broader capabilities, such as iterative self-correction, flexible generation orders (e.g., infilling) and a more generalized modeling framework~\cite{yu2025discrete,ji2025denoising,ye2024diffusion}. 
This trend has recently culminated in hybrid architectures like block-wise diffusion, which fuse a global autoregressive structure with local parallel diffusion, preserving causal flow for variable-length generation while benefiting from intra-block parallelism~\citep{arriola2025block, cheng2025sdar,wang2025diffusion,wu2025fastv2}.

\subsection{Multimodal Large Language Diffusion Models}
Inspired by successes in language, discrete diffusion has been extended to the multimodal domain, first targeting understanding tasks. Pioneering works like LLaDA~\cite{zhu2025llada}, LaViDA~\cite{li2025lavida}, and Dimple~\cite{yu2025dimple} established the paradigm: image and text features are concatenated and denoised as a single sequence via a global masked diffusion process. 
This foundation was quickly built upon by more ambitious efforts toward a unified architecture for both understanding and generation, such as Lumina-DiMOO~\cite{xin2025lumina}, MMaDA~\cite{yang2025mmada}, Lavida-O~\cite{li2025lavidao} and Muddit~\cite{shi2025muddit}, which frame all tasks within a single, elegant denoising framework. 

However, this rapid pursuit of unification has exposed a foundational weakness. These models, whether for understanding or generation, predominantly rely on a global diffusion process. This approach suffers from crippling quadratic complexity and a notoriously difficult learning objective when applied to long, mixed-modality sequences, often resulting in ``suboptimal performance" on complex reasoning tasks~\cite{you2025llada, xin2025lumina}. We argue that \textbf{a robust and efficient understanding model is a prerequisite for any unified system}. Consequently, we are the first to systematically apply the block-wise diffusion paradigm~\cite{arriola2025block, cheng2025sdar} to this problem. Our approach circumvents the scalability issues of global diffusion by combining intra-block parallel denoising with an efficient inter-block autoregressive structure, creating a more powerful and practical foundation for multimodal understanding.

\section{Theoretical Analysis of SDAR-VL}
\label{sec:theory}

In this section, we formalize the training dynamics of BD3 and SDAR-VL and show that:
(i) asynchronous block-wise noise scheduling strictly reduces the variance of the stochastic gradient estimator (thus stabilizing training), and
(ii) the standard BD3 loss scaling based on the sampled noise ratio $t_b$ is a biased estimate of an ideal NELBO objective, while our effective mask ratio $t'_b$ yields an unbiased estimator.

Throughout, we focus on a single block $b$ within a sequence and treat the contribution of that block to the loss or to a particular gradient coordinate as a scalar random variable~$Z$. All statements therefore apply coordinate-wise to the full gradient vector. For notational simplicity, we omit the explicit dependence on $x$ and~$\theta$ whenever it is clear from context.

\subsection{Theoretical Setup and Notation}
\label{subsec:theory-setup}

For a given training step, consider a sequence $x$ decomposed into $B$ blocks, and a noise level $t$ sampled from a scheduling distribution $\mathcal P(\cdot\mid\tau)$. Given $t$, a binary mask $\bm m \in \{0,1\}^{L'}$ of block length $L'$ is sampled (e.g., via independent Bernoulli or more structured masking), inducing the \emph{realized} or \emph{effective} mask ratio
{\small
\begin{equation}
 t' = \frac{\|\bm m\|_1}{L'}.
\end{equation}
}
For block $b$, we denote the corresponding block-level loss by
{\small
\begin{equation}
  \ell_b
  =
  \sum_{\ell \in \mathcal{M}_{t_b}^b}
  - \log p_\theta\!\big(x_0^{b,\ell} \mid x_{t_b}^b, x^{<b}\big),
\end{equation}
}
where $\mathcal{M}_{t_b}^b$ is the set of masked positions in block $b$ under the mask $\bm m^b$.

To analyze variance and bias, we abstract the contribution of block $b$ at a given step into a scalar random variable
{\small
\begin{equation}
  Z = Z(t_b, \bm m^b),
\end{equation}
}
which can be either the block loss itself $Z = \ell_b$ or any fixed linear functional of the gradient, e.g.\ $Z = u^\top \nabla_\theta \ell_b$ for some fixed vector $u$. For a given noise level $t$, define
{\small
\begin{equation}
  \mu(t) := \E[Z \mid t],
  \qquad
  \sigma^2(t) := \Var(Z \mid t).
\end{equation}
}
Expectations are taken over data, masks, and any other randomness conditioned on~$t$.

\subsection{Variance reduction via asynchronous block-wise noise scheduling}
\label{subsec:theory-async-variance}

We compare two ways of sampling the noise levels for a batch of $B$ blocks:

\begin{itemize}
  \item \textbf{Synchronous schedule:} A single noise level $t$ is sampled from $\mathcal P(\cdot\mid\tau)$ and shared by all $B$ blocks in the sequence. Conditioned on $t$, we obtain i.i.d.\ block contributions
  $Z_1,\dots,Z_B \stackrel{\text{i.i.d.}}{\sim} Z \mid t$, and the batch-level estimator is
  {\small
  \begin{equation}
    \bar Z_{\mathrm{sync}}
    =
    \frac{1}{B} \sum_{b=1}^B Z_b.
  \end{equation}
  }
  \item \textbf{Asynchronous block-wise schedule:} Each block $b$ independently samples its own noise level $t_b \sim \mathcal P(\cdot\mid\tau)$ and mask~$\bm m^b$. Thus, $(t_b, Z_b)$ are i.i.d.\ across blocks, and the batch-level estimator is
  {\small
  \begin{equation}
    \bar Z_{\mathrm{async}}
    =
    \frac{1}{B} \sum_{b=1}^B Z_b.
  \end{equation}
  }
\end{itemize}

Note that the same notation $Z_b$ is used in both cases, but in the synchronous setting all $Z_b$ share the same $t$, whereas in the asynchronous setting each $Z_b$ is associated with an independently drawn~$t_b$.

\paragraph{Lemma 1 (Unchanged expectation).}
Under the two sampling schemes above, the estimators $\bar Z_{\mathrm{sync}}$ and $\bar Z_{\mathrm{async}}$ have the same expectation:
{\small
\begin{equation}
  \E[\bar Z_{\mathrm{sync}}]
  =
  \E[\bar Z_{\mathrm{async}}]
  =
  \E[Z].
  \label{eq:async-expectation}
\end{equation}
}

\emph{Proof.}
By the law of total expectation,
{\small
\begin{equation}
  \E[\bar Z_{\mathrm{sync}}]
  = \E_{t}\big[\E[\bar Z_{\mathrm{sync}} \mid t]\big]
  = \E_t[\mu(t)]
  = \E[Z].
\end{equation}
}
In the asynchronous setting, $(Z_b)_{b=1}^B$ are i.i.d.\ copies of $Z$, so
{\small
\begin{equation}
  \E[\bar Z_{\mathrm{async}}]
  = \E\Big[\frac{1}{B} \sum_{b=1}^B Z_b\Big]
  = \E[Z].
\end{equation}
}

\paragraph{Lemma 2 (Variance reduction).}
Assume $B \ge 2$ and that the conditional mean $\mu(t)$ is not almost surely constant as a function of $t$, i.e.\ $\Var_t(\mu(t)) > 0$. Then
{\small
\begin{align}
  \Var(\bar Z_{\mathrm{async}})
  &= \frac{1}{B}
    \Big(
      \E_t[\sigma^2(t)] + \Var_t(\mu(t))
    \Big),
  \label{eq:var-async} \\
  \Var(\bar Z_{\mathrm{sync}})
  &= \Var_t(\mu(t))
    + \frac{1}{B} \E_t[\sigma^2(t)].
  \label{eq:var-sync}
\end{align}
}
Subtracting \eqref{eq:var-async} from \eqref{eq:var-sync} gives
{\small
\begin{equation}
  \Var(\bar Z_{\mathrm{sync}})
  - \Var(\bar Z_{\mathrm{async}})
  =
  \Big(1 - \tfrac{1}{B}\Big)\,
  \Var_t(\mu(t)).
  \label{eq:var-diff}
\end{equation}
}
This quantity is strictly positive whenever $B>1$ and $\Var_t(\mu(t))>0$.

\emph{Proof.}
For the synchronous scheme, we apply the law of total variance:
{\small
\begin{equation}
  \Var(\bar Z_{\mathrm{sync}})
  = \Var\!\big(\E[\bar Z_{\mathrm{sync}}\mid t]\big)
    + \E\big[\Var(\bar Z_{\mathrm{sync}}\mid t)\big].
\end{equation}
}
Conditioned on $t$, the $Z_b$ are i.i.d.\ with conditional mean $\mu(t)$ and variance $\sigma^2(t)$, whence
{\small
\begin{equation}
  \E[\bar Z_{\mathrm{sync}} \mid t] = \mu(t),
  \qquad
  \Var(\bar Z_{\mathrm{sync}} \mid t)
  = \frac{1}{B} \sigma^2(t).
\end{equation}
}
Substituting gives \eqref{eq:var-sync}. In the asynchronous scheme, $(Z_b)_{b=1}^B$ are i.i.d.\ draws from the marginal distribution of $Z$, so
{\small
\begin{equation}
  \Var(\bar Z_{\mathrm{async}})
  = \frac{1}{B} \Var(Z).
\end{equation}
}
Again by the law of total variance,
{\small
\begin{equation}
  \Var(Z)
  = \E_t[\sigma^2(t)] + \Var_t(\mu(t)),
\end{equation}
}
which yields \eqref{eq:var-async}. The difference \eqref{eq:var-diff} then follows directly.
\hfill$\square$

\paragraph{Remark 1 (Training stability).}
Lemma~2 shows that for any fixed noise distribution $\mathcal P(\cdot\mid\tau)$ and model parameters $\theta$, asynchronous block-wise noise scheduling leaves the expectation of the batch estimator unchanged (Lemma~1) while reducing its variance by $(1 - 1/B)\Var_t(\mu(t))$. Empirically, we observe a strong dependence of the loss on the noise level $t$, which implies that $\Var_t(\mu(t))$ is non-negligible. In this regime, asynchronous scheduling stabilizes training by averaging over multiple noise levels \emph{within} a single step, rather than concentrating the variability induced by different $t$ across different steps.

\subsection{Bias of standard loss scaling and unbiasedness of the effective mask ratio}
\label{subsec:theory-bias}

We now formalize the bias introduced by the standard BD3 loss scaling using the sampled ratio $t_b$, and show that replacing $t_b$ with the realized mask ratio $t'_b$ restores an unbiased estimate of a natural NELBO objective.

\paragraph{Ideal objective based on realized corruption.}
Intuitively, the ``strength'' of supervision contributed by block $b$ should be proportional to the \emph{actual} fraction of tokens that are masked, rather than the nominal target~$t_b$. This motivates the following ideal objective:
{\small
\begin{equation}
  \Lstar(\theta)
  :=
  \E\Big[
    -\frac{\ell_b}{t'_b}
  \Big],
  \qquad
  t'_b = \frac{\|\bm m^b\|_1}{L'}.
  \label{eq:ideal-objective}
\end{equation}
}
Here $t_b$ is drawn from the scheduling distribution $\mathcal P(\cdot\mid\tau)$, and $\bm m^b$ is sampled conditioned on $t_b$.

\paragraph{Standard BD3 scaling as a biased estimator.}
In the standard BD3 framework, the block-wise contribution is scaled by the \emph{sampled} noise ratio~$t_b$:
{\small
\begin{equation}
  V := -\frac{\ell_b}{t_b},
  \qquad
  \Lorig(\theta) := \E[V].
\end{equation}
}
Comparing $\Lorig(\theta)$ with the ideal objective $\Lstar(\theta)$ in \eqref{eq:ideal-objective} gives a precise expression for the bias.

\paragraph{Lemma 3 (Bias of standard BD3 scaling).}
The standard BD3 objective $\Lorig(\theta)$ is in general a biased approximation of $\Lstar(\theta)$:
{\small
\begin{equation}
  \Lorig(\theta) - \Lstar(\theta)
  =
  \E\Big[
    -\ell_b
    \Big(
      \frac{1}{t_b} - \frac{1}{t'_b}
    \Big)
  \Big].
  \label{eq:bias-diff}
\end{equation}
}
Unless $t_b = t'_b$ almost surely, the right-hand side is non-zero whenever $\ell_b$ correlates with the realized corruption level.

\emph{Proof.}
By definition,
{\small
\begin{equation}
  \Lorig(\theta)
  =
  \E\Big[-\frac{\ell_b}{t_b}\Big],
  \qquad
  \Lstar(\theta)
  =
  \E\Big[-\frac{\ell_b}{t'_b}\Big].
\end{equation}
}
Subtracting these gives \eqref{eq:bias-diff}. If $t_b \ne t'_b$ with positive probability and the loss $\ell_b$ depends monotonically on the realized corruption level $t'_b$, then $\ell_b$ is non-trivially correlated with $\tfrac{1}{t_b} - \tfrac{1}{t'_b}$, and the expectation in \eqref{eq:bias-diff} does not vanish.

It is sometimes useful to decompose the bias in terms of mean and covariance:
{\small
\begin{align}
  \Lorig(\theta) - \Lstar(\theta)
  &=
  \E[\ell_b]\;
  \E\Big[\frac{1}{t'_b} - \frac{1}{t_b}\Big] \nonumber \\
  &\quad+
  \mathrm{Cov}\Big(
    \ell_b,
    \frac{1}{t_b} - \frac{1}{t'_b}
  \Big),
  \label{eq:bias-cov}
\end{align}
}
highlighting that the mismatch between $t_b$ and $t'_b$ distorts the relative weighting of easier and harder blocks.

\paragraph{A simple approximation illustrating the dependence on block size.}
To obtain further intuition, consider the following stylized approximation. Suppose that, for a fixed block length $L'$, the block loss can be expressed as
{\small
\begin{equation}
  \ell_b
  \approx
  L' \, t'_b \, h(t'_b),
\end{equation}
}
where $h(t')$ denotes the average negative log-likelihood per masked token at realized mask ratio $t'$. Empirically, $h$ is non-decreasing and often convex in $t'$: higher corruption ratios lead to more difficult reconstructions. Assume that the mask is sampled via independent Bernoulli($t_b$) trials, so that
{\small
\begin{equation}
  t'_b = t_b + \varepsilon,
  \qquad
  \E[\varepsilon \mid t_b] = 0,
  \qquad
  \Var(\varepsilon \mid t_b)
  = \frac{t_b(1 - t_b)}{L'}.
\end{equation}
}
In this approximation, the difference between the two block contributions becomes
{\small
\begin{equation}
  V - U
  :=
  -\frac{\ell_b}{t_b}
  +
  \frac{\ell_b}{t'_b}
  \;\approx\;
  -L' h(t'_b)
  \Big(\frac{t'_b}{t_b} - 1\Big),
  \quad
  U := -\frac{\ell_b}{t'_b}.
\end{equation}
}
A second-order Taylor expansion of $h(t'_b)$ around $t_b$ and taking conditional expectations over $\varepsilon$ yields a bias term that scales proportionally to $\Var(\varepsilon\mid t_b) \propto 1/L'$. Thus, the discrepancy between $t_b$ and $t'_b$---and hence the bias in \eqref{eq:bias-diff}---is most pronounced for small blocks (small $L'$) or high-variance masking schemes.

\paragraph{Effective mask ratio as an unbiased estimator.}
Our proposed effective mask ratio replaces $t_b$ with the realized ratio $t'_b$ in the scaling factor:
{\small
\begin{equation}
  \Lau(\theta)
  :=
  \E\Big[
    -\frac{\ell_b}{t'_b}
  \Big].
\end{equation}
}

\paragraph{Lemma 4 (Unbiasedness of effective mask ratio).}
For any parameter $\theta$,
{\small
\begin{equation}
  \Lau(\theta)
  =
  \Lstar(\theta).
\end{equation}
}
Under standard regularity assumptions (e.g., dominated convergence),
{\small
\begin{equation}
  \nabla_\theta \Lau(\theta)
  = \E\Big[ -\,\tfrac{\nabla_\theta \ell_b}{t'_b} \Big],
  \qquad
  \nabla_\theta \Lau(\theta)
  = \nabla_\theta \Lstar(\theta).
\end{equation}
}
In other words, using $t'_b$ in the normalization yields an unbiased estimator of the ideal objective \eqref{eq:ideal-objective} and its gradient.

\emph{Proof.}
By definition, $\Lau(\theta) = \E[-\ell_b / t'_b]$, which coincides with $\Lstar(\theta)$ in \eqref{eq:ideal-objective}. Under the stated regularity conditions, differentiation and expectation can be interchanged, giving
{\small
\begin{equation}
  \nabla_\theta \Lau(\theta)
  =
  \E\Big[
    -\frac{\nabla_\theta \ell_b}{t'_b}
  \Big]
  =
  \nabla_\theta \Lstar(\theta).
\end{equation}
}

\paragraph{Remark 2 (Interaction with asynchronous scheduling).}
Combining Lemma~2 and Lemma~4, SDAR-VL can be viewed as optimizing the ideal objective $\Lstar(\theta)$ using:
(i) an unbiased, variance-reduced gradient estimator achieved by asynchronous block-wise noise sampling, and
(ii) an unbiased scaling of each block's contribution based on its realized corruption level $t'_b$. Together, these modifications remove a systematic mis-scaling present in the standard BD3 objective and reduce gradient variance, leading to more stable training dynamics in practice.

\section{Experiments}
\label{appendix:experiment}
\subsection{Training Setting}
\label{app:training_settings}
\paragraph{Datasets and Training Stages}

\begin{table*}[htbp]
\centering
\caption{\textbf{Training Hyperparameters and Datasets.} We detail the configurations for each sequential training stage. M-SI and M-OV denote the single-image and one-vision subsets of MAmmoTH~\cite{guo2024mammoth}, while VW represents VisualWebInstruct~\cite{jia2025visualwebinstruct}. Our model is trained on the first five datasets sequentially. The final dataset, LLaVA-NeXT~\cite{liu2024llavanext}, is used for full fine-tuning in our ablation studies (Sec.~\ref{sec:ablations}). \textbf{The presented settings are applied consistently for training both the 4B and 8B versions of our model.}
}
\resizebox{0.95\linewidth}{!}{
\label{tab:training_settings}
\renewcommand{\arraystretch}{1.2}
\begin{tabular}{l|c|c|c|c|c|c|c}
\toprule
\textbf{Training Data} & \textbf{LLaVA-Pretrain} & \textbf{M-SI} & \textbf{M-OV} & \textbf{VW} & \textbf{M-OV+VW} & \textbf{R1-Onevision}&\textbf{LLaVA-NeXT} \\
\midrule
Vision Tower & \multicolumn{7}{c}{SigLIP-2-So400m-patch14-384 \cite{tschannen2025siglip}} \\
Language Tower & \multicolumn{7}{c}{{SDAR-Chat (4B and 8B)}~\cite{cheng2025sdar}} \\
Attention Mechanism & \multicolumn{7}{c}{Blockwise Attention} \\
\hline
Batch Size & 32 & 64 & 64 & 64 & 64 & 32 & 32 \\
Packing Length & 8192 & 16384 & 16384 & 16384 & 16384 & 16384 & 8192 \\
\# Samples & 558K & 10M & 2M & 900K & 3M &155K& 738K \\
Epochs & 2 & 1 & 1 & 1 & 1 & 3 & 1 \\
Training Tokens & 1.1B & 33.8B & 9.3B & 1.8B & 11.1B & 2.6B & 2.7B \\
\hline
LR of Vision Tower & --- & $2 \times 10^{-6}$ & $2 \times 10^{-6}$ & $2 \times 10^{-6}$ & $2 \times 10^{-6}$ & $2 \times 10^{-6}$ & $2 \times 10^{-6}$ \\
LR of Language Tower & --- & $1 \times 10^{-5}$ & $1 \times 10^{-5}$ & $1 \times 10^{-5}$ & $1 \times 10^{-5}$ & $1 \times 10^{-5}$ & $1 \times 10^{-5}$ \\
LR of Projector & $1 \times 10^{-3}$ & $1 \times 10^{-5}$ & $1 \times 10^{-5}$ & $1 \times 10^{-5}$ & $1 \times 10^{-5}$ & $1 \times 10^{-5}$ & $1 \times 10^{-5}$\\
\bottomrule
\end{tabular}
}
\end{table*}
Our training curriculum follows a sequential, multi-stage approach designed to progressively build the model's capabilities. 
The main training pipeline consists of six steps, which are organized into four stages, as detailed below. 
The specific hyperparameters for each step are provided in Appendix~\ref{app:training_settings}, Table~\ref{tab:training_settings}.

\begin{itemize}
\item \textbf{Stage 1: Vision-Language Alignment.} We begin by pre-training on the {LLaVA-Pretrain} dataset~\cite{liu2023visual}. This initial stage is dedicated to aligning the projector, connecting the visual features from the vision tower to the language model's embedding space.

\item \textbf{Stage 2: Capability Expansion.} In the second stage, we fine-tune the model on the comprehensive MAmmoTH-VL dataset~\cite{guo2024mammoth}. This is done sequentially, first on {M-SI} (10M single-image samples) and then on {M-OV} (2M diverse-modality samples), to equip the model with a broad understanding of multimodal concepts.

\item \textbf{Stage 3: Reasoning Enhancement.} This stage focuses on enhancing complex reasoning. We first fine-tune on the reasoning-centric {VisualWebInstruct (VW)} dataset~\cite{jia2025visualwebinstruct}. To further cultivate balanced reasoning skills, we conclude training with a mixture of {M-OV and VW} data.

\item \textbf{Stage 4: Long CoT Distillation.} We further distill long-form chain-of-thought reasoning by fine-tuning \textit{SDAR-VL-Instruct} on \textit{R1-OneVision}~\cite{yang2025r1}. 
This step injects multi-step reasoning patterns and yields \textit{SDAR-VL-Think}, which improves performance on challenging math benchmarks.

\end{itemize}

\subsection{Evaluation Setting}
\label{app:eval_settings}
\paragraph{Evaluation Protocol}
For SDAR-VL, we adopt block-inference-based low-confidence greedy static decoding, with both the block length and the number of denoising steps fixed to 4. The decoding procedure follows SDAR~\cite{cheng2025sdar}. Unless otherwise specified, all reported results for SDAR-VL use this configuration.

For the open-source baselines in Table~\ref{tab:image-bench}, including the InternVL series, Qwen-VL series, and LLaVA-OneVision-OV models, we run local evaluations using the VLMEval-Kit framework~\cite{duan2025vlmevalkitopensourcetoolkitevaluating} with standard greedy decoding and the default evaluation configurations provided by the toolkit. The multi-image and video results in Table~\ref{tab:multi-video-bench} follow the same evaluation setup and sources as Table~\ref{tab:image-bench}.
For closed-source models, the GPT-4V and GPT-4o results are directly taken from the LLaVA-OneVision paper~\cite{li2024llava}. For discrete diffusion baselines in Table~\ref{tab:image-bench}, the LLaDA-V scores are from its original paper~\cite{you2025llada}, while the remaining diffusion-based results are taken from LaViDA-O~\cite{li2025lavidao}. For the reasoning-focused comparison in Table~\ref{tab:reasoning-model}, we use the Qwen2.5-VL and R1-OneVision-7B results reported in R1-OneVision~\cite{yang2025r1}, and the LaViDA-8B results reported in LaViDA~\cite{li2025lavida}.

\paragraph{Evaluation Benchmarks}
To comprehensively assess the capabilities of our SDAR-VL models (4B and 8B), we conduct an extensive evaluation across a diverse suite of 21 established benchmarks. These benchmarks are carefully selected to probe a wide spectrum of multimodal abilities, from foundational visual perception to complex, high-level reasoning. The evaluation suite is organized into four key categories:

\begin{itemize}
\item \textbf{General Visual Understanding \& Reliability:} We assess core vision-language capabilities and factual correctness using established benchmarks, including MMBench~\cite{liu2024mmbench}, SeedBench~\cite{li2023seed}, MME~\cite{fu2025mmecomprehensiveevaluationbenchmark}, MM-Vet~\cite{yu2023mm}, and MMStar~\cite{chen2024we}. 
To assess knowledge-intensive understanding, we evaluate on MMMU~\cite{yue2024mmmu}.
In addition, we evaluate robustness against object hallucination using HallusionBench~\cite{guan2024hallusionbench}.

\item \textbf{Mathematical and Scientific Reasoning:} We test advanced reasoning skills on benchmarks requiring complex problem-solving. This includes MathVista~\cite{lu2023mathvista}, MathVerse~\cite{zhang2024mathverse} (for which we use chain-of-thought (CoT) decoding; in Table~\ref{tab:image-bench}, “vo” and “vd” denote the vision-only and vision-dominant settings, respectively), and MathVision~\cite{wang2024measuring} for mathematical reasoning in visual contexts, and ScienceQA~\cite{lu2022learn} for multimodal scientific question answering.

\item \textbf{Document-centric VQA:} We evaluate the model's ability to understand and reason over text-rich images across diverse formats using DocVQA~\cite{mathew2021docvqa}, ChartQA~\cite{masry2022chartqa}, InfoVQA~\cite{mathew2022infographicvqa}, and AI2D~\cite{kembhavi2016diagram}.

\item \textbf{Multi-image and Video Understanding:} We evaluate multi-image reasoning with MuirBench~\cite{wang2024muirbench} and BLINK~\cite{fu2024blink}. For video understanding, we use MVBench~\cite{li2024mvbench}, VideoMME~\cite{fu2025video}, and MLVU~\cite{zhou2024mlvu} to assess temporal comprehension and cross-frame aggregation.
\end{itemize}

\section{Example Appendix}

\begin{casebox}[HallusionBench]
\small

\begin{center}
    \includegraphics[width=0.8\linewidth]{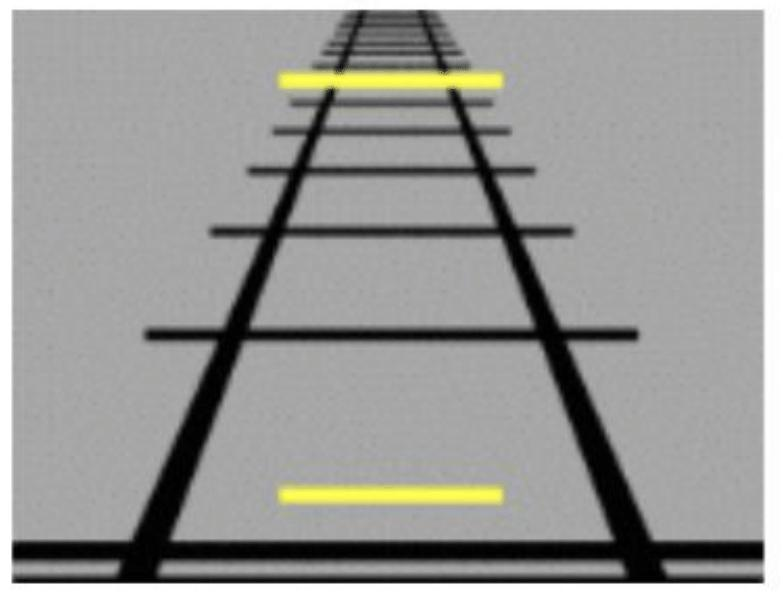}
\end{center}

\vspace{0.6em}
\textbf{Question:}
Are the two yellow lines the same length?

\vspace{0.4em}
\textbf{Ground-truth answer:}
Yes

\vspace{0.4em}
\textbf{Model output (SDAR-VL-Instruct-8B):}

\noindent
To determine if the two yellow lines are the same length, we need to analyze the image and compare the lengths of the yellow lines.

1. \textbf{Identify the Yellow Lines:}

    The first yellow line is located near the top of the image.

    The second yellow line is located near the bottom of the image.

2. \textbf{Compare the Lengths:}

    Visually inspect the lengths of both yellow lines.\\
    Note that both yellow lines appear to span the same horizontal distance across the image.

3. \textbf{Conclusion:} 

    Based on the visual comparison, it can be concluded that the two yellow lines are indeed the same length.

Therefore, the answer is yes.

\normalsize
\end{casebox}

\begin{casebox}[MME: artwork]
\small

\begin{center}
    \includegraphics[width=0.95\linewidth]{figures/cases/11331.jpg}
\end{center}

\vspace{0.6em}
\textbf{Question:}
Is this artwork created by donatello? Please answer yes or no.

\vspace{0.4em}
\textbf{Ground-truth answer:}
Yes

\vspace{0.4em}
\textbf{Model output (SDAR-VL-Instruct-8B):}

\noindent
To determine if the artwork in the image was created by Donatello, we need to analyze the style and characteristics of the artwork and compare them to Donatello's known works.

1. \textbf{Identify the Artwork}: The image shows a detailed relief sculpture with figures in dynamic poses, which is characteristic of Renaissance art.

2. \textbf{Donatello's Style}: Donatello was an Italian sculptor of the early Renaissance period, known for his detailed and expressive sculptures. His works often feature human figures with a high degree of realism and emotional expression.

3. \textbf{Comparison to Known Works}: Donatello's famous works include "David," "St. George," and "The Feast of Herod." These sculptures are known for their intricate details, dynamic compositions, and the use of perspective.

4. \textbf{Analysis of the Image}: The relief sculpture in the image exhibits similar characteristics to Donatello's works:
   
   The figures are depicted with a high level of detail and realism.
   The composition is dynamic, with figures arranged in a way that suggests movement and interaction.
   
   The use of perspective and the three-dimensional of the figures are indicative of Donatello's style.

5. \textbf{Conclusion}: Based on the stylistic similarities between the artwork in the image and Donatello's known works, it is reasonable to conclude that this artwork was indeed created by Donatello.

Therefore, the answer is: Yes.

\normalsize
\end{casebox}

\begin{casebox}[MMMU: Financial Accounting]
\small

\begin{center}
    \includegraphics[width=0.95\linewidth]{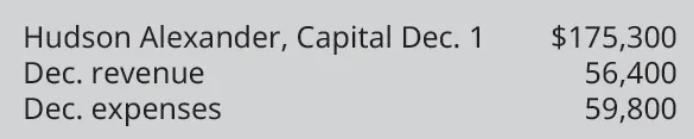}
\end{center}

\vspace{0.6em}
\textbf{Question:}
Here are facts for the Hudson Roofing Company for December. <image 1> Assuming no investments or withdrawals, what is the ending balance in the owners' capital account?

A \$171,900	 \quad B \$170,000	\quad C \$172,500\quad D \$181,900
\vspace{0.4em}
\textbf{Ground-truth answer:}
Yes

\vspace{0.4em}
\textbf{Model output (SDAR-VL-Instruct-8B):}

\noindent
{To determine the ending balance in the owners' capital account, we need to consider the initial capital, the revenue, and the expenses for the month of December.}

1. \textbf{Initial Capital}: {The starting capital is } \$175,300.

2. \textbf{Revenue}: {The revenue for December is } \$56,400.

3. \textbf{Expenses}: {The expenses for December are } \$59,800.

{To find the ending balance, we add the revenue to the initial capital and then subtract the expenses:}
$
\text{Ending Balance} = \text{Initial Capital} + \text{Revenue} - \text{Expenses}
$

$
\text{Ending Balance} = 175,300 + 56,400 - 59,800
$

$
\text{Ending Balance} = 171,900
$

{Therefore, the ending balance in the owners' capital account is } \$171,900. {The correct answer is A.}

\normalsize
\end{casebox}

\begin{casebox}[MMbench: finegrained\_perception]
\small

\begin{center}
    \includegraphics[width=0.95\linewidth]{figures/cases/1846.jpg}
\end{center}

\vspace{0.6em}
\textbf{Question:}
What direction is Czechia in the Mediterranean Sea?

A. east \quad B. south \quad C. west \quad D. north 

\vspace{0.4em}
\textbf{Ground-truth answer:}
D

\vspace{0.4em}
\textbf{Model output (SDAR-VL-Instruct-8B):}

\noindent
To determine the direction of Czechia relative to the Mediterranean Sea, we need to analyze the geographical positions of both locations on the map.

1. \textbf{Identify the Mediterranean Sea}: The Mediterranean Sea is located in the southern part of Europe, bordered by countries such as Spain, France, Italy, Greece, and others.

2. \textbf{Locate Czechia}: Czechia is situated in Central Europe, bordered by countries such as Poland, Slovakia, Austria, Germany, and the Czech Republic.

3. \textbf{Determine the Direction}: 
   
   The Mediterranean Sea is to the south of Czechia.
   
   Therefore, Czechia is to the north of the Mediterranean Sea.

Given these observations, the correct answer is:

D. north

\normalsize
\end{casebox}

\begin{casebox}[SeedBench: instance reasoning]
\small

\begin{center}
    \includegraphics[width=0.95\linewidth]{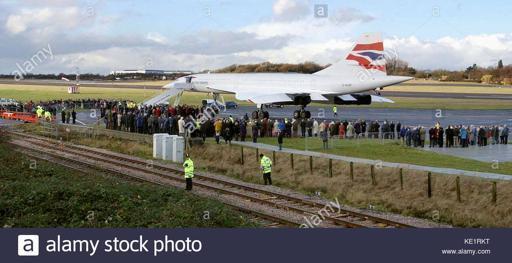}
\end{center}

\vspace{0.6em}
\textbf{Question:}
Which object is closest to the crowd in the image?

A Airliner	 \quad B TrainTrain	\quad C Jet\quad D None of the above
			
\vspace{0.4em}
\textbf{Ground-truth answer:}
A

\vspace{0.4em}
\textbf{Model output (SDAR-VL-Instruct-8B):}

\noindent
To solve the problem, I will analyze the image to identify the object closest to the crowd. I will describe the image, focusing on the relevant details, and then use logical reasoning to determine the correct answer from the given options. 

 The image shows a large crowd of people gathered on a grassy area near a runway. A large airliner is visible on the runway, with its nose pointing towards the crowd. There is also a train track adjacent to the crowd, but a train is not visible in the image. 

 The question asks which object is closest to the crowd. The airliner is positioned on the runway, close to the crowd. The train track is next to the crowd, but no train is present. Therefore, the airliner is the object closest to the crowd. 

 A

\normalsize
\end{casebox}

\begin{casebox}[Mathverse: geometry (angle)]
\small

\begin{center}
    \includegraphics[width=0.95\linewidth]{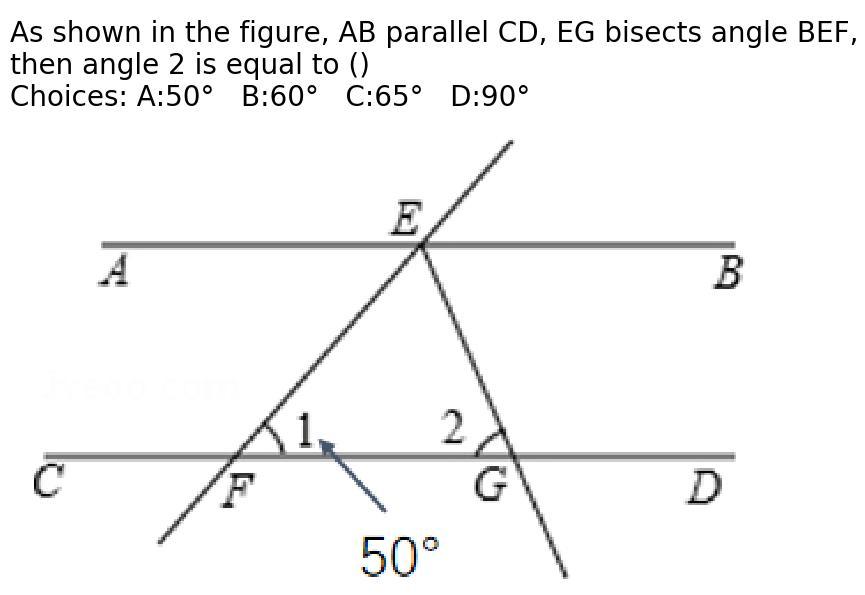}
\end{center}

\vspace{0.6em}
\textbf{Question:}
According to the question shown in the image, first provide your reasoning,
then give the measure of angle~2 and the correct option letter (A--D).

\vspace{0.4em}
\textbf{Ground-truth answer:}
65$^\circ$ (option C).

\vspace{0.4em}
\textbf{Model output (SDAR-VL-Think-8B):}

\noindent\texttt{<think>}

Okay, so I'm trying to solve this geometry problem. Let me analyze the image directly.
The image shows two parallel lines, AB and CD, with a transversal EG intersecting them.
Line EG bisects angle BEF, and there's an angle marked as 50 degrees at point F.
The question asks for the measure of angle~2, which is equal to angle EGF.

First, let's focus on the parallel lines AB and CD. Since EG is a transversal, it creates
corresponding angles and alternate interior angles. The image shows that EG bisects angle
BEF, meaning it splits angle BEF into two equal parts.

Given that angle F is 50 degrees, and considering the properties of parallel lines,
angle BEF must be supplementary to angle F. Therefore, angle BEF is 180 degrees minus
50 degrees, which equals 130 degrees. Since EG bisects angle BEF, each of the two angles
created by the bisector will be half of 130 degrees, i.e., 65 degrees.

This means angles BEG and FEG are both 65 degrees. Now, consider angle EGF: since EG
is a transversal and bisects angle BEF, angle EGF is one of the angles created by the
bisector. Therefore, angle EGF is 65 degrees. Thus, angle~2, which is equal to angle
EGF, is also 65 degrees.

\noindent\texttt{</think>}

The measure of angle~2 is 65 degrees, which corresponds to option~C. Answer: C.

\normalsize
\end{casebox}

\begin{casebox}[Mathverse: analytic geometry]
\small
\label{case:hyperbola-2185}

\begin{center}
    \includegraphics[width=0.6\linewidth]{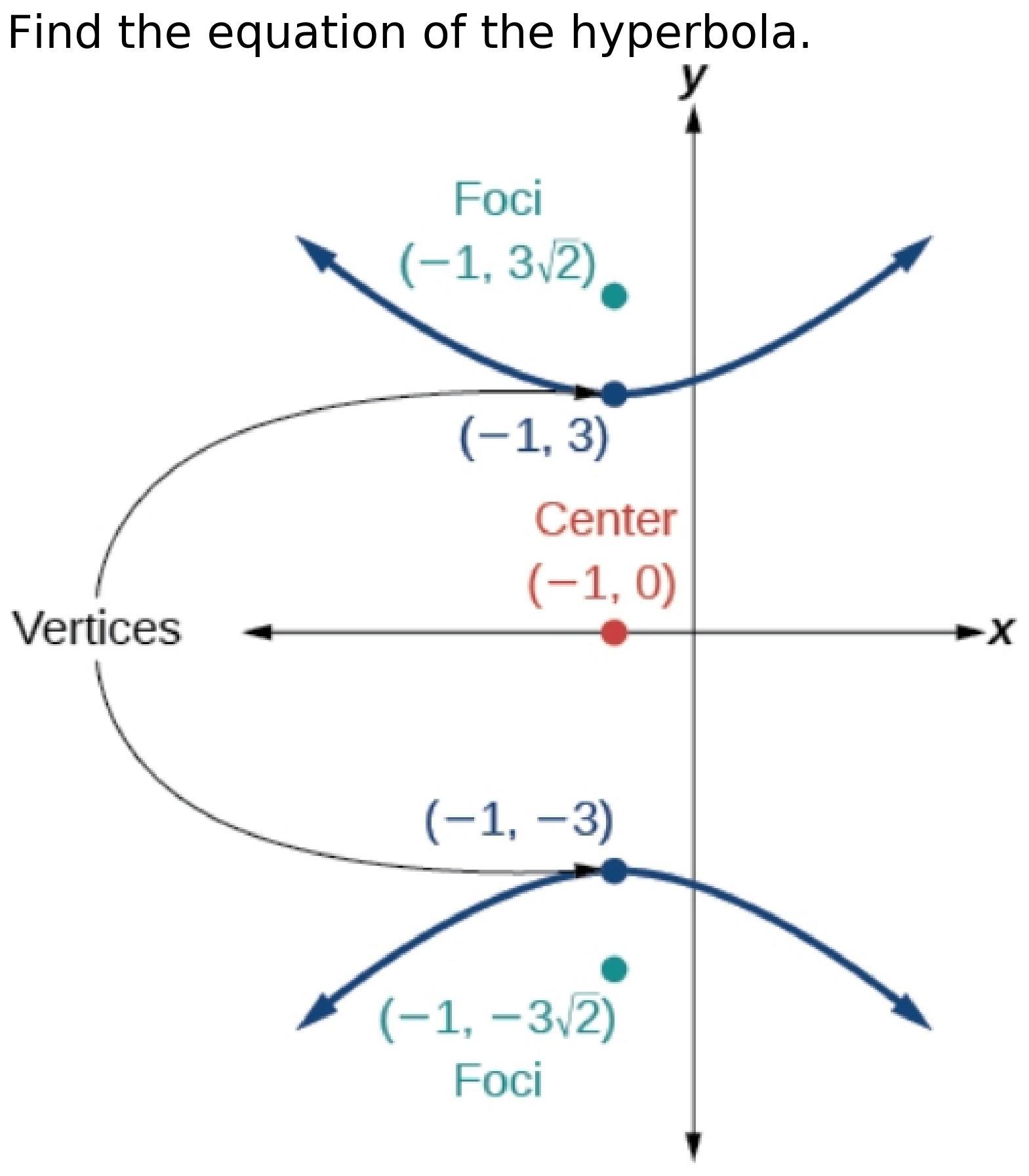}
\end{center}

\vspace{0.6em}
\textbf{Question:}
Find the equation of the hyperbola.

\vspace{0.4em}
\textbf{Ground-truth answer:}
$\displaystyle \frac{y^2}{9} - \frac{(x + 1)^2}{9} = 1.$

\vspace{0.4em}
\textbf{Model output (SDAR-VL-Think-8B):}

\noindent\texttt{<think>}

Okay, so I need to figure out the equation of the hyperbola based on the image provided.
Let me start by recalling what I know about hyperbolas. A hyperbola is a type of conic
section that has two separate branches. The standard form of its equation depends on
whether it's oriented horizontally or vertically.

As seen in the image, the center is at $(-1, 0)$, which means the hyperbola is shifted
to the left by 1 unit. The center is also the midpoint between the vertices, which are
at $(-1, 3)$ and $(-1, -3)$. Since both vertices have the same $x$-coordinate, they lie
vertically above and below the center. This indicates that the hyperbola is vertical,
meaning its transverse axis is along the $y$-axis.

For a vertical hyperbola, the standard form of the equation is
$\dfrac{(y - k)^2}{a^2} - \dfrac{(x - h)^2}{b^2} = 1$, where $(h, k)$ is the center.
In this case, $h = -1$ and $k = 0$, so the equation becomes
$\dfrac{y^2}{a^2} - \dfrac{(x + 1)^2}{b^2} = 1$.

Next, I need to determine the values of $a$ and $b$. The distance from the center to
each vertex is $a$. Since the vertices are at $(-1, 3)$ and $(-1, -3)$, the distance
from the center $(-1, 0)$ to each vertex is 3 units. Therefore, $a = 3$, so $a^2 = 9$.

The foci of the hyperbola are at $(-1, 3\sqrt{2})$ and $(-1, -3\sqrt{2})$. The distance
from the center to each focus is $c = 3\sqrt{2}$. For a hyperbola, the relationship
between $a$, $b$, and $c$ is $c^2 = a^2 + b^2$. Plugging in the known values gives
$c^2 = 18$ and $a^2 = 9$, so $b^2 = 18 - 9 = 9$.

Putting everything together, the equation becomes
$\dfrac{y^2}{9} - \dfrac{(x + 1)^2}{9} = 1$.

\noindent\texttt{</think>}

\vspace{0.4em}
\textbf{Step-by-step summary.}
(1) Identify the center at $(-1, 0)$.
(2) Use the vertical vertices $(-1, \pm 3)$ to get $a^2 = 9$.
(3) Use the foci $(-1, \pm 3\sqrt{2})$ to get $c^2 = 18$ and hence $b^2 = 9$.
(4) Substitute into the standard form of a vertical hyperbola.

\textbf{Answer:} $\displaystyle \frac{y^2}{9} - \frac{(x + 1)^2}{9} = 1$.
\normalsize
\end{casebox}

\begin{casebox}[Mathvision: metric geometry ]
\small
\label{case:hyperbola-2185}

\begin{center}
    \includegraphics[width=0.4\linewidth]{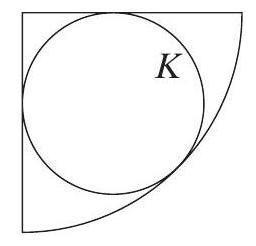}
\end{center}

\vspace{0.6em}
\textbf{Question:}
Question: A circle $K$ is inscribed in a quarter circle with radius 6 as shown in the figure. What is the radius of circle $K$?

\vspace{0.4em}

(A) $\frac{6-\sqrt{2}}{2}$
(B) $\frac{3 \sqrt{2}}{2}$
(C) 2.5
(D) 3
(E) $6(\sqrt{2}-1)$

\vspace{0.4em}
\textbf{Ground-truth answer:}
E

\vspace{0.4em}
\textbf{Model output (SDAR-VL-Think-8B):}

\noindent\texttt{<think>}\\
Okay, so I have this geometry problem involving a circle inscribed in a quarter circle. The quarter circle has a radius of 6, and I need to find the radius of the smaller circle $ K $.

First, let's consider the image itself. The image shows a quarter circle with a radius of 6, and inside it, there's a smaller circle $ K $ that is tangent to both the arc and the two sides of the quarter circle. This means the smaller circle $ K $ is touching the curved part of the quarter circle and the two straight sides.

Next, let's focus on the positioning of the smaller circle $ K $. As seen in the image, the center of the smaller circle $ K $ is at a distance equal to its radius from both sides of the quarter circle. If we place the quarter circle in a coordinate system with its center at the origin (0,0), the two sides of the quarter circle would be along the x-axis and y-axis. The center of the smaller circle $ K $ would then be at some point (r, r), where r is the radius of the smaller circle.

Now, let's consider the distance from the center of the smaller circle $ K $ to the arc of the quarter circle. This distance is the radius of the quarter circle minus the radius of the smaller circle, which is 6 - r. However, this distance is also the hypotenuse of a right triangle with legs of length r each. Therefore, we can use the Pythagorean theorem to express this distance as $ \sqrt{r^2 + r^2} = \sqrt{2r^2} = r\sqrt{2} $.

Setting these two expressions for the distance equal to each other, we get:
$$ 6 - r = r\sqrt{2} $$

Solving for r, we rearrange the equation:
$$ 6 = r + r\sqrt{2} $$
$$ 6 = r(1 + \sqrt{2}) $$
$$ r = \frac{6}{1 + \sqrt{2}} $$

To rationalize the denominator, we multiply the numerator and the denominator by the conjugate of the denominator:
$$ r = \frac{6(1 - \sqrt{2})}{(1 + \sqrt{2})(1 - \sqrt{2})} $$
$$ r = \frac{6(1 - \sqrt{2})}{1 - 2} $$
$$ r = \frac{6(1 - \sqrt{2})}{-1} $$
$$ r = -6(1 - \sqrt{2}) $$
$$ r = 6(\sqrt{2} - 1) $$

Thus, the radius of the smaller circle $ K $ is $ 6(\sqrt{2} - 1) $.

Answer: E\\
\noindent\texttt{</think>}

\vspace{0.4em}
The radius of the smaller circle inscribed in the quarter circle is calculated using geometric relationships and the Pythagorean theorem. By setting up an equation based on the distances from the center of the smaller circle to the sides and the arc, we determine that the radius is $ 6(\sqrt{2} - 1) $.

\textbf{Answer:} E

\normalsize
\end{casebox}

\end{document}